\documentclass[10pt,twocolumn,letterpaper]{article}

\usepackage[accsupp]{axessibility}

\usepackage{subcaption} 
\usepackage{graphbox}

\usepackage{wacv}
\usepackage{times} 
\usepackage{epsfig}
\usepackage{graphicx}
\usepackage{amsmath}
\usepackage{amssymb}
\usepackage{booktabs}

\DeclareMathOperator*{\argmax}{argmax}
\DeclareMathOperator*{\argmin}{argmin}
\usepackage{multicol}
\usepackage{multirow} 
\usepackage{mathtools} 
\DeclarePairedDelimiterX\set[1]\lbrace\rbrace{#1}

\usepackage{float}

\def\wacvPaperID{641} 

\wacvalgorithmstrack   

\wacvfinalcopy 

\ifwacvfinal
\usepackage[breaklinks=true,bookmarks=false]{hyperref}
\else
\usepackage[pagebackref=true,breaklinks=true,colorlinks,bookmarks=false]{hyperref}
\fi

\begin{document}

\title{Closer Look at the Transferability of Adversarial Examples: \\
How They Fool Different Models Differently}

\author{Futa Waseda$^1$, Sosuke Nishikawa$^1$, Trung-Nghia Le$^{2,3,4}$, Huy H. Nguyen$^{2}$, and Isao Echizen$^{1,2}$ \\
\small{$^1$The University of Tokyo, Tokyo, Japan\ \ \ \ \ \ \ \ \ \ \ \ \ \ \ $^2$National Institute of Informatics, Tokyo, Japan} \\
\small{$^3$University of Science, VNU-HCM, Vietnam\ \ \ \ \ \ \ \ \ \ \ \ \ \ \ $^4$Vietnam National University, Ho Chi Minh City, Vietnam} \\
{\tt\small futa-waseda@g.ecc.u-tokyo.ac.jp; \{nhhuy,iechizen\}@nii.ac.jp}
}

\maketitle
\thispagestyle{empty}

\begin{abstract}
Deep neural networks are vulnerable to adversarial examples (AEs), which have adversarial transferability: AEs generated for the source model can mislead another (target) model's predictions. However, the transferability has not been understood in terms of to which class target model's predictions were misled (i.e., class-aware transferability). In this paper, we differentiate the cases in which a target model predicts the same wrong class as the source model (``same mistake") or a different wrong class (``different mistake") to analyze and provide an explanation of the mechanism. We find that (1) AEs tend to cause same mistakes, which correlates with ``non-targeted transferability"; however, (2) different mistakes occur even between similar models, regardless of the perturbation size. Furthermore, we present evidence that the difference between same mistakes and different mistakes can be explained by non-robust features, predictive but human-uninterpretable patterns: different mistakes occur when non-robust features in AEs are used differently by models. Non-robust features can thus provide consistent explanations for the class-aware transferability of AEs.
\end{abstract}

\section{Introduction}
\label{sec:introduction}

\begin{figure}[t]
    \centering
        \includegraphics[width=\linewidth]{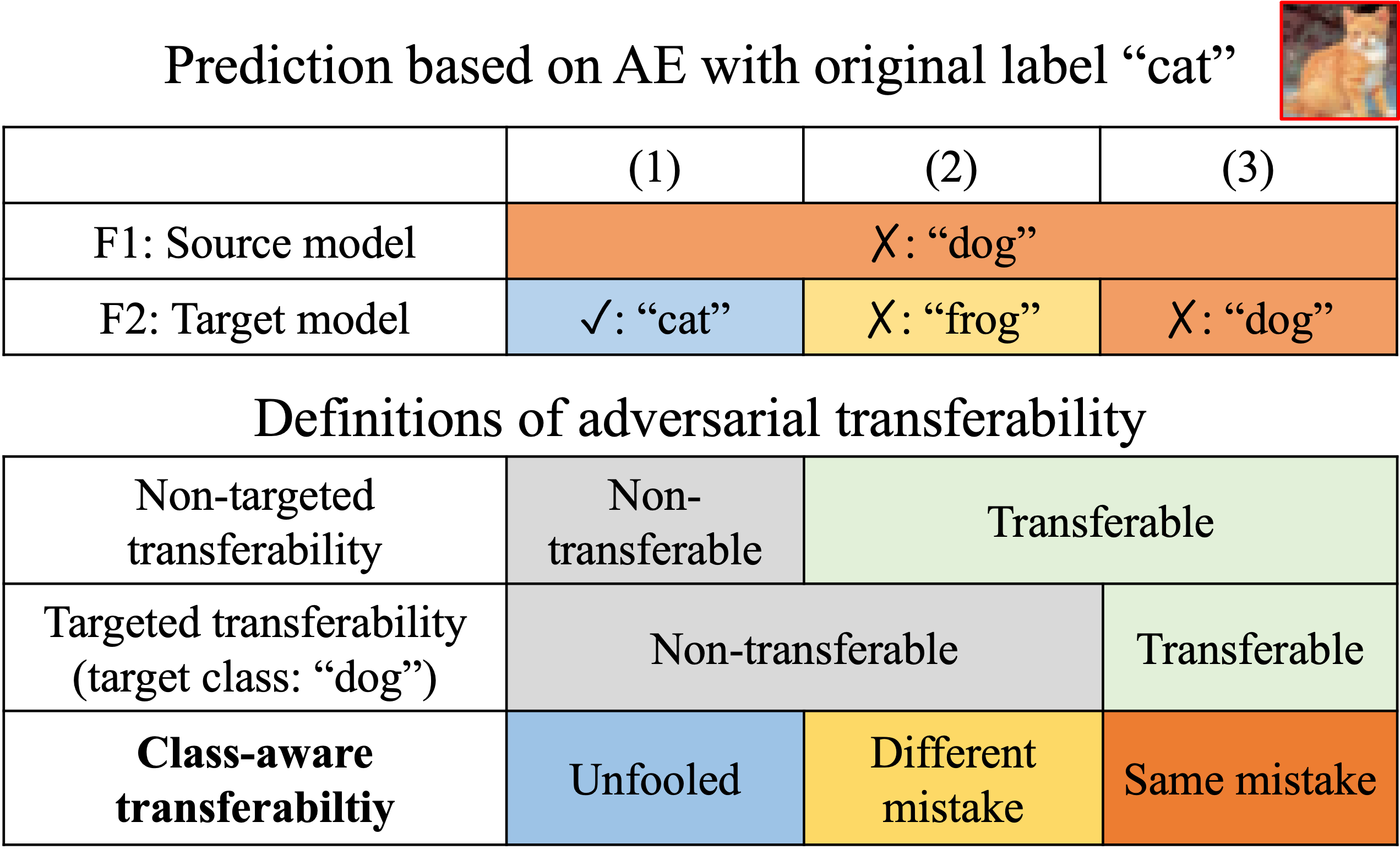} 
        \caption{Example case of classifying transferability of adversarial example (AE). When the source model misclassifies the AE as ``dog", class-aware transferability of the AE to the target model is classified as (1) the target model correctly classified the AE as ``cat" (unfooled), (2) it misclassified the AE as ``frog" (different mistake), or (3) it misclassified the AE as ``dog” (same mistake). 
        In contrast, non-targeted and targeted transferability are defined in a binary way: whether the AE met the attackers' objective or not.}
        \label{fig:classify_transferability}
\end{figure}

\begin{figure}[t]
    \centering
        \includegraphics[width=\linewidth]{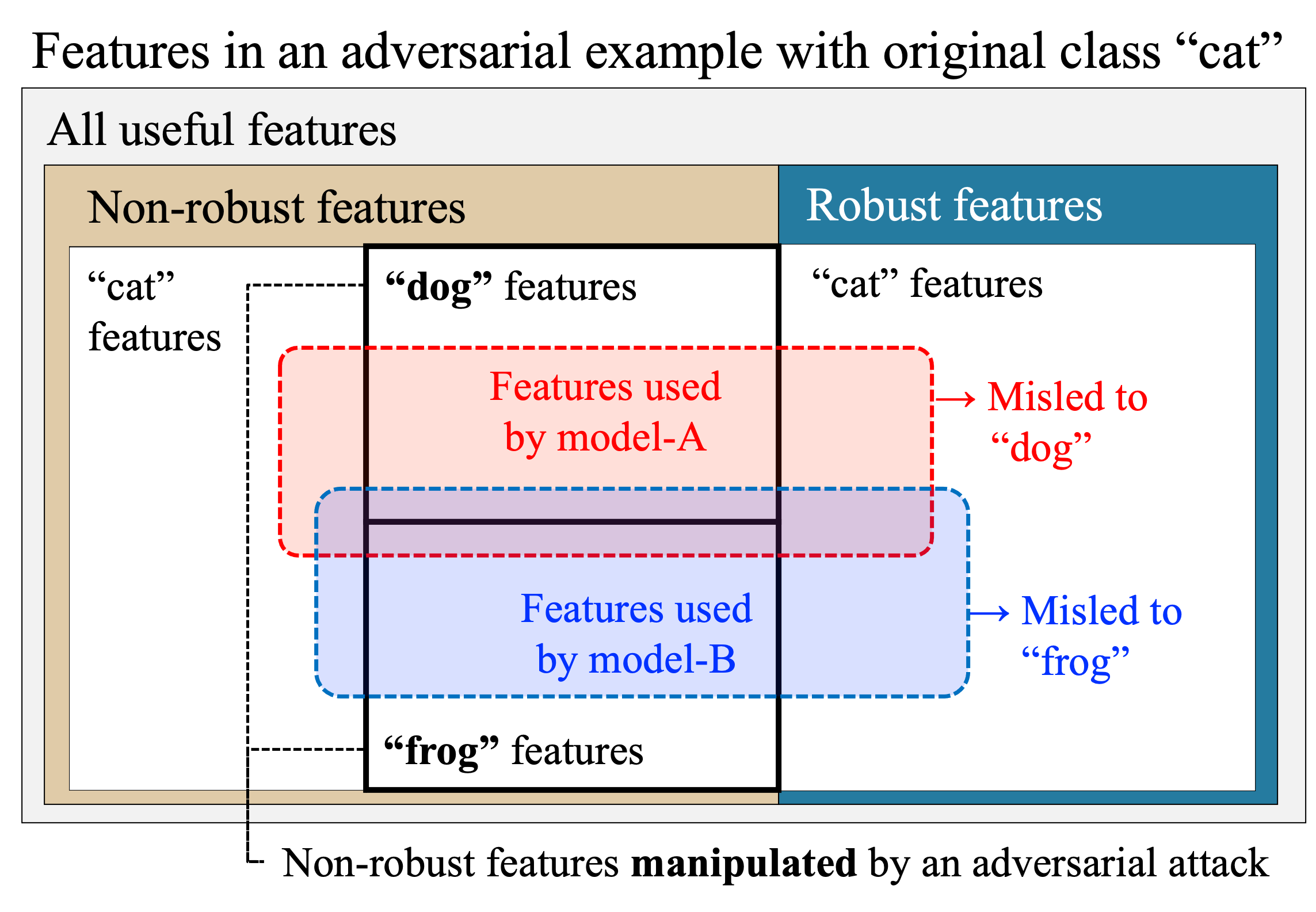} 
        \caption{Our hypothesis for how AEs cause different models to make different predictions: non-robust features \cite{ilyas2019AE_non_rob} can be used differently by models. 
        Let us say a ``cat" image is manipulated by an adversarial attack, and a part of non-robust features correlates with ``dog" class (``dog" features), and another part correlates with ``frog" class (``frog" features). 
        When model-A uses more of the ``dog" features than the ``frog" features, and model-B does the opposite, model-A may predict the AE as ``dog" and model-B may predict the AE as ``frog." 
        Our work shows that non-robust features can cause both ``different mistakes" and ``same mistakes."
        }
        \label{fig:hypothesis}
\end{figure}

Deep neural networks (DNNs) are vulnerable to adversarial examples (AEs), which are slightly perturbed by noises or patterns to mislead DNNs' predictions \cite{szegedy2013intriguing, goodfellow2014explaining_harnessing_ae}. 
Since AEs can fool DNNs without affecting human perception, they are a severe threat to real-world DNN applications, even in the physical world \cite{kurakin2016adversarial_BIM}. 
However, existing defensive techniques remain vulnerable to AEs due to a lack of understanding of adversarial vulnerability. 
A critical property of AEs that needs to be better understood is their transferability: AEs generated using the source model may also fool other models \cite{szegedy2013intriguing, goodfellow2014explaining_harnessing_ae, papernot2016transferability_ml_black_box, papernot2017practical_blackbox}. 
This transferability allows attackers to use a substitute model to generate AEs to fool other unknown (target) models with different architectures or weights (i.e., a ``black-box attack" \cite{papernot2016transferability_ml_black_box}), which poses a considerable risk in our society. 
Understanding the transferability is essential to reducing the risk of black-box attacks and understanding the fundamental problem in current DNNs that cause adversarial vulnerability.

Many studies \cite{szegedy2013intriguing, goodfellow2014explaining_harnessing_ae, papernot2016transferability_ml_black_box,  dong2019benchmarking, liu2016delving_into_trans, demontis2019adversarial} investigate (1) non-targeted and (2) targeted transferability, depending on the objective of adversarial attacks.
Non-targeted transferability is defined for non-targeted attacks, which aim to fool a model regardless of the misclassified class; on the other hand, targeted transferability is defined for targeted attacks, which aim to fool a model towards a specific target class (illustrated in Figure~\ref{fig:classify_transferability}).
Primarily, previous works focused on explaining the non-targeted transferability \cite{goodfellow2014explaining_harnessing_ae, liu2016delving_into_trans, tramer2017space_of_trans, ilyas2019AE_non_rob}: they showed that similarity between the source and target model allows AEs to fool them simultaneously.
However, it is unclear to which class the source and target models' predictions are misled, which we refer to as ``class-aware transferability."
Although the cases where different models misclassify an AE to the same class (``same mistake") and different classes (``different mistake") are different phenomena, we do not know what factors affect their proportion and what their mechanisms are.


With this motivation, we analyze the transferability from a novel perspective of ``class-aware transferability." 
We aim to understand the transferability phenomenon rather than simple risk evaluation.
First, we perform a detailed analysis of the factors that affect class-aware transferability.
We then tested whether our hypothesis explains the observed transferability phenomenon.

We analyze class-aware transferability under two conditions: model similarity and perturbation size. 
Class-aware transferability differentiates the cases where the target model misclassifies the AE as the same class as the source model (``same mistake") and a different class than the source model (``different mistake") (Figure~\ref{fig:classify_transferability}).
We present three main findings: (1) AEs tend to cause same mistakes, which is strongly connected to their capability of fooling target models (non-targeted transferability); (2) AEs cause different mistakes even on very similar models; (3) larger perturbations intend to cause same mistakes but do not reduce the ratio of different mistakes.

To provide a comprehensive explanation of the mechanisms AEs causing different mistakes and same mistakes, we provide a novel insight based on the theory of ``non-robust feature."
Ilyas et al. \cite{ilyas2019AE_non_rob} showed that an AE may have predictive but human-imperceptible features (i.e., non-robust features) of the class to which a model was misled.
Here, same mistakes are the logical consequence based on their theory; however, they do not explain different mistakes.  
In this work, we show that AEs that cause different mistakes can have the non-robust features of the two different classes to which the two models were misled. 
It indicates that the dependency of models on the learning features can cause different mistakes: when the non-robust features in AEs are used differently by different models, those models may classify the AEs differently (Figure~\ref{fig:hypothesis}). 
As we strengthen the theory of non-robust features \cite{ilyas2019AE_non_rob}, we support the claims that AEs are at least partly a consequence of learning ``superficial cues" \cite{jo2017measuring_sur_statis_bengio} or ``shortcuts" \cite{geirhos2020shortcut}. 

Our contributions are summarized as follows.
\begin{itemize}
    \item Our evaluation of class-aware transferability shows (1) that AEs tend to cause same mistakes, which is strongly connected to their  capability of fooling target models (non-targeted transferability), (2) that different mistakes occur even between source and target models with high similarity, and (3) that larger perturbations do not reduce different mistakes, indicating a misalignment in misleading the source and target model towards the same class.
    
    \item We provide an explanation of the mechanisms causing different and same mistakes by extending the theory of non-robust features. Same mistakes are due to AEs having the non-robust features of the class to which the model was misled. When the manipulated non-robust features in the AEs are used differently by different models, those models may classify the AE differently.
    
\end{itemize}


\section{Related Work}
\label{sec:related_work}

\subsection{Non-targeted Adversarial Transferability} 
Non-targeted adversarial transferability is defined by whether or not the target model assigns a wrong class rather than the true (original) class. 
Szegedy et al. \cite{szegedy2013intriguing} showed that AEs transfer even when the source and target models have different architectures or are trained on a disjoint dataset. 
Papernot et al. \cite{papernot2016transferability_ml_black_box} showed that non-targeted AEs transfer even between different machine learning methods such as DNNs, SVMs, and decision trees. 
Naseer et al. \cite{naseer2019cross_domain_trans} generated AEs that transfer even between models trained on different image domains, such as cartoon and painting. 
Although these studies show intriguing transferability, how such AEs affect the target model's predictions is unclear. 
This paper analyzes class-aware transferability, differentiating different and same mistakes.

The transferability of non-targeted adversarial attacks has been explained by the similarity between the source and target models. 
Goodfellow et al. \cite{goodfellow2014explaining_harnessing_ae} showed that adversarial perturbations are highly aligned with the weight vectors of a model and that different models learn similar functions when trained on the same dataset to perform the same task. 
Liu et al. \cite{liu2016delving_into_trans} revealed by visualization that transferability can arise from the similarity of the decision boundary that separates the true class and other classes. 
Tramer et al. \cite{tramer2017space_of_trans} asserted that transferability appears when ``adversarial subspaces" intersect between different classifiers. 
Ilyas et al. \cite{ilyas2019AE_non_rob} showed that adversarial vulnerability can arise from non-robust features that are predictive but uninterpretable by humans and that transferability arises from the similarity of learned non-robust features between models. 
However, these do not clarify when and why different or same mistakes occur. 
We are the first to provide insightful explanations and discussions of their mechanisms based on the theory of non-robust features.

\subsection{Targeted Adversarial Transferability} 
Targeted adversarial transferability is defined by whether or not the target model assigns the same class as the target class towards which the source model was attacked. 
Liu et al. \cite{liu2016delving_into_trans} showed that, in contrast to non-targeted attacks, targeted attacks rarely transfer between models. 
Class-aware transferability allows us to directly compare the effect of non-targeted and targeted AEs, instead of using two different metrics of non-targeted and targeted transferability.

Several studies improved the transferability of targeted attacks by a similar idea: avoiding overfitting to the image or source model. 
Dong et al. \cite{dong2018boosting_trans_momentum} used momentum in iterations of a gradient-based adversarial attack; Xie et al. \cite{xie2019improv_trans_input_diversity} increased input diversity when generating AEs, and Nasser et al. \cite{naseer2021generating_trans_tar} generated class-specific AEs by using a generative adversarial network (GAN) to capture the global data distribution rather than overfitting the source model and the single image. 
However, these efforts did not provide a theoretical explanation of the mechanism causing same mistakes. 
A few studies explained same mistakes.
Goodfellow et al. \cite{goodfellow2014explaining_harnessing_ae} hypothesized that the linear behavior of neural networks explains it. Such behavior is acquired by generalizing to solve the same task, thus resembling a linear classifier trained on the same data. 
Ilyas et al. \cite{ilyas2019AE_non_rob} provided a widely accepted explanation: models can assign the same class by looking at similar non-robust features in AEs. 
However, these do not explain our observation that different mistakes occur between similar models regardless of the perturbation size. 
This paper provides a novel insight based on the theory of non-robust features to explain both different and same mistakes.

\subsection{Adversarial Examples causing Different Predictions}
Several works have studied how AEs cause different models to make different predictions, which corresponds to the cases of unfooled or different mistakes.
Nakkiran et al. \cite{nakkiran2019discussion_just_bugs} generated AEs that only fool the source model and do not fool another model with the same architecture and trained on the same dataset. They claim that there exist AEs that exploit directions irrelevant to the true data distribution and thus irrelevant to features. 
Tramer et al. \cite{tramer2017space_of_trans} used MNIST data with XOR artifacts to train linear and quadratic models and generated AEs that fooled only either. They hypothesized that AEs might not transfer when two models learn different features. 
Charles et al. \cite{charles2019geometric_pers} discussed from a geometric perspective and illustrated the decision boundaries and directions of the gradients when AEs fool only a linear classifier but not a two-layer ReLU classifier. 
Our hypothesis for how AEs cause different models to make different predictions can largely explain these cases and provide further interpretations. 

\subsection{Class-wise Robustness}
Some works focused on class-wise robustness, which evaluates robustness for each class separately.
A few works revealed that the class-wise robustness of models trained by adversarial training (AT) \cite{madry2017towards_pgd_at_adversarial_training} is imbalanced, which can be interpreted by our non-robust features hypothesis (Figure~\ref{fig:hypothesis}).
AT is a defense method that trains models incorporating AEs into training data.
Tian et al. \cite{tian2021analysis_class_wise_rob} revealed the imbalance in class-wise robustness of AT models and its fluctuation during training.
Xia et al. \cite{xia2021class_aware_rt_cart} showed that the robustness of a specific vulnerable class improves by using the AEs weighted for that vulnerable class in AT.
These findings are interpreted by our findings as follows: AT tries to force a model to ignore non-robust features in AEs. Therefore, the class-wise robustness in AT depends on which class of the non-robust features AEs contain, and its balance between classes can be a critical factor in determining class-wise robustness.

\section{Adversarial transferability analysis}
\label{sec:analysis}


\subsection{Overview}

In this section, we evaluate the class-aware transferability of AEs by differentiating ``different mistakes" and ``same mistakes." 
We aim to clarify the factors that affect class-aware transferability. 
Firstly, we analyze the effect of model factors by gradually changing the similarity between the source and target models. 
Different from Liu et al. \cite{liu2016delving_into_trans}, we not only compare models with different architectures but also with different or the same initial weights and models that are only in different training epochs.  
In addition, we use the metric of the decision boundary distance defined by Tramer et al. \cite{tramer2017space_of_trans} as a quantitative model similarity measurement.
Secondly, we evaluate class-aware transferability by gradually increasing the perturbation size. 

\subsubsection{Class-aware Transferability} 
\label{subsec:class_aware}
We classify transferability by whether the target model was ``unfooled,” whether it made a ``different mistake,” or a ``same mistake.” 
The term ``same mistake" was mentioned by Liu et al. \cite{liu2016delving_into_trans} and was not the focus of their study. 

The focus of our study is to evaluate how the malicious effect of AEs generated for a source model $F1$ can affect the classification results of an (unknown) target model $F2$. 
Therefore, we evaluate the transferability only for the AEs generated for the original images correctly classified by both $F1$ and $F2$ and  successfully fooled $F1$:
\small
\begin{gather}
 (x',y,y1) \sim D'_{F1,F2} = \left\{ (x, y)\sim D \;
 \begin{tabular}{|l}
 $F1(x)=y$, \\ 
 $F2(x)=y$, \\
 $F1(x')=y1 \;(\neq y)$. \\
 \end{tabular}
 \right\}
\end{gather}
\normalsize
where an AE $x'=adv(x, y, F1)$ is generated by an adversarial attack $adv(\cdot)$ for the image-label pair $(x,y)$ in the original set $D$, and $y1 (\neq y)$ denotes the wrong class that the source model misclassified. For these AEs, we define the metrics for class-aware transferability as follows.
\begin{enumerate}
 \item Unfooled ratio: \small $\mathbf{P}_{(x', y, y1) \sim D'_{F1,F2}} \left[F2(x')=y\right]$ \normalsize
 \item Fooled ratio: \small $\mathbf{P}_{(x', y, y1) \sim D'_{F1,F2}} \left[F2(x')\neq y\right]$ \normalsize
    \begin{enumerate}
        \item Different mistake ratio:
        \small
             \begin{equation}
                \mathbf{P}_{(x', y, y1) \sim D'_{F1,F2}} \left[F2(x')=y2 \right], where \; y2 \notin \{y,y1\}
            \end{equation}
        \normalsize
        \item Same mistake ratio: 
        \small
            \begin{equation}
                \mathbf{P}_{(x', y, y1) \sim D'_{F1,F2}} \left[F2(x')=y1 \right]
            \end{equation}
        \normalsize
    \end{enumerate}
\end{enumerate}

If the target model $F2$ classifies an AE $x'$ as the true class $y$, it is unfooled; if it classifies the AE as a different wrong class $y2$ than the source model $F1$, it makes a different mistake; if it classifies the AE as the same wrong class $y1$ as the source model $F1$, it makes a same mistake. 

Note that fooled ratio corresponds to non-targeted transferability. 
Same mistake ratio corresponds to targeted transferability only if $y1$ is the target class of a targeted attack.

\subsubsection{Generation of Adversarial Examples} 

We examine both non-targeted attacks, which aim to fool a model regardless of the misclassified class, and targeted attacks, which aim to fool a model towards a specific target class $y^{tar}$. The optimization problems are formulated as
\small
\begin{eqnarray}
     \text{(Non-targeted:)}\;\; \argmax_{x'} L(x', y)\\
     \text{(Targeted:)}\;\; \argmin_{x'} L(x', y^{tar})
\end{eqnarray}
\normalsize
where $L(\cdot)$ is a loss function and $x'$ is the AE generated from the original input $x$. Both are subject to an $l_p$-bound,
\begin{math}
    \| x' - x \|_p < \epsilon
\end{math}, 
so that $x'$ remains sufficiently close to $x$.

We generate AEs using two gradient-based attacks: (1) the fast gradient method (FGM), which is an efficient method to generate $l_p$ bounded AEs (the generalized version of the fast gradient sign method \cite{goodfellow2014explaining_harnessing_ae}), and (2) the projected gradient descent (PGD) method \cite{madry2017towards_pgd_at_adversarial_training}, which is the iterative version of FGM that generates stronger AEs. 
We provide results for other attacks, such as MIM \cite{dong2018boosting_trans_momentum}, CW \cite{carlini2017towards_cw_attack}, and DeepFool \cite{moosavi2016deepfool}, in supplementary material.

\subsubsection{Measurement of Model Similarity}
\label{subsec:measure_trans}

For quantitative measurement of the similarity between the source and target models, we use a method devised by Tramer et al. \cite{tramer2017space_of_trans}.
It measures the average distance of the decision boundary for $N$ images between two models:
\small
\begin{equation}
\label{eq:Dist}
    Dist(F1, F2) = \frac{1}{N}\sum_{i=1}^N |d(F1, x_i) - d(F2, x_i)|
\end{equation}
\normalsize
where 
\begin{math}
    d(f,x)=\argmin_\epsilon [f(x+\epsilon \cdot v) \neq y]
\end{math}
is the minimum distance from an input $x$ to the decision boundary of model $f$. The distance is calculated in the direction of the vector 
\begin{math}
    v = {\nabla_{x} L(x, y; F1)}/{\| \nabla_{x} L(x, y; F1) \|_2}
\end{math}, 
which is a normalized vector of the non-targeted adversarial perturbation generated for the source model $F1$. Therefore, this metric is directly related to the non-targeted transferability. 
We use this metric to analyze the relationship between class-aware transferability and model similarity indicated by non-targeted transferability.
To calculate the equation~\ref{eq:Dist}, we randomly chose 1,000 images from the test set that all models correctly classified.


\subsection{Evaluation Settings}
\label{subsection: eval}

\subsubsection{Dataset} 
We used Fashion-MNIST \cite{xiao2017fashion_mnist}, CIFAR-10 \cite{krizhevsky2009learning_cifar10} and STL-10 \cite{coates2011analysis_stl10} datasets, which are all ten-class datasets.
We generated AEs $l_2$-bounded by a specific $\epsilon$ (assuming that the pixels take values in the range [0, 1]). 
The PGD attack iterates for ten steps with step size $\alpha$=$\epsilon/5$. 
To generate targeted AEs, we randomly choose target classes for each image.
For a fair comparison, we evaluated 2,000 random images from the test set that all models correctly classified.

\subsubsection{Models} 
For Fashion-MNIST, we examined models with four simple architectures: fully-connected networks with 2 or 4 hidden layers (FC-2 or FC-4) and convolutional networks with 2 or 4 convolution layers followed by two fully-connected layers (Conv-2 or Conv-4).
For CIFAR-10 and STL-10, we examined models with five popular architectures: VGG-16, VGG-19 \cite{simonyan2014very_vgg}, ResNet-18, ResNet-34 \cite{he2016deep_resnet}, and DenseNet-121 \cite{huang2017densely}. 
We trained all models for 40 epochs for Fashion-MNIST, and 100 epochs for CIFAR-10 and STL-10 (details in supplementary material). 
For precise analysis, we independently trained three models for each architecture: two models trained using the same initial weight parameters and one trained using the other initial weights (when the initial weights are the same between models with the same architecture, the only difference is the randomness of the shuffled training data or dropout layers). 
In addition, we also compare early versions of the source model as target models at the $i^{th}$ epoch. 
Hereinafter, models $F2$ with ``(w:same)" or ``(w:diff)" in their name are the models independently trained using the same or different initial weights as used for F1; ``(v:$i$)" is F1 at the $i^{th}$ epoch.

\subsection{Results and Discussions}

\begin{figure*}[t]
    \centering
        \begin{subfigure}[align=t]{.325\textwidth}
          \centering
          \includegraphics[width=1\linewidth]{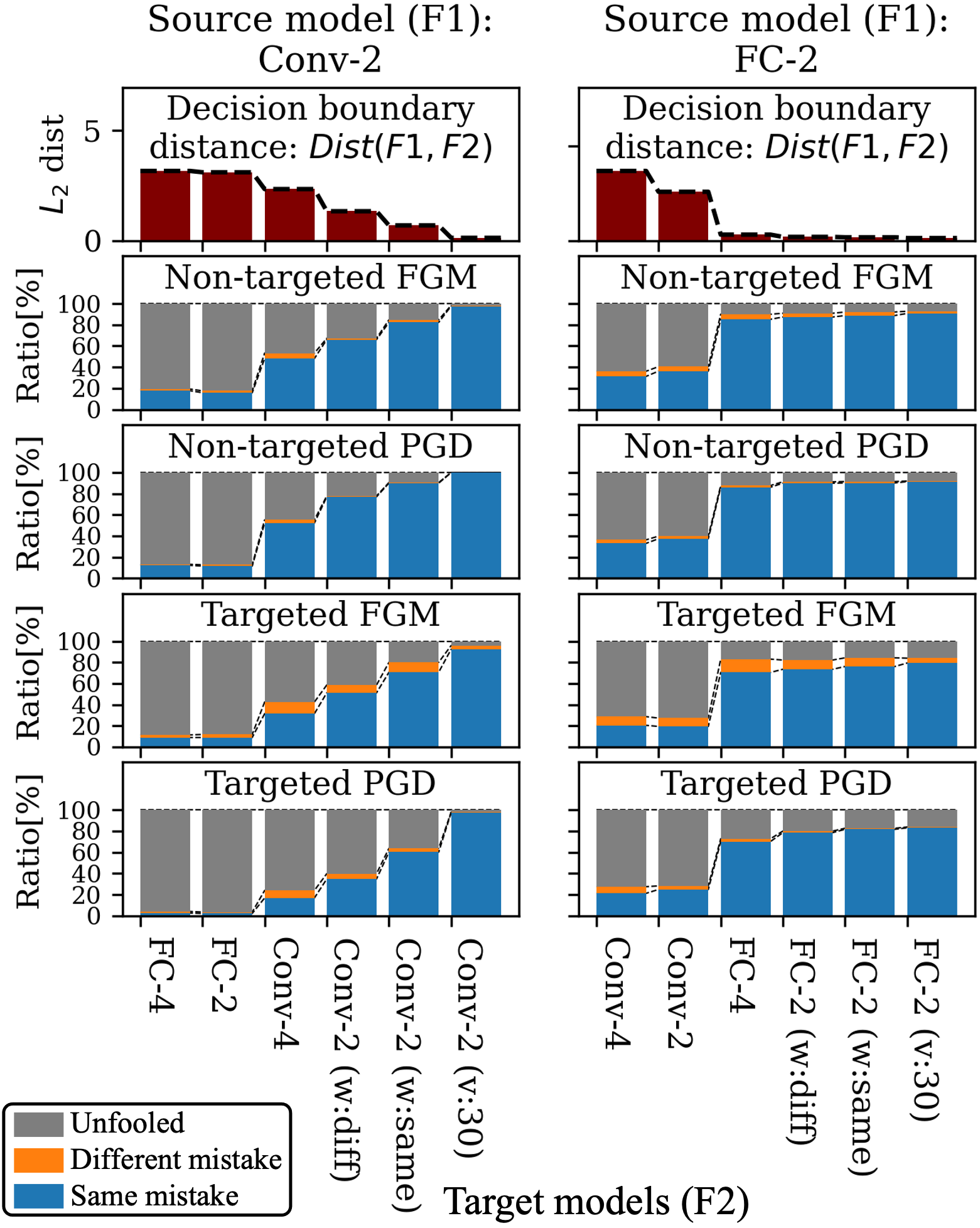}
          \caption[9pt,align=t]{Fashion-MNIST ($\epsilon$=1.0)}
          \label{fig:analysis_fashion_mnist}
        \end{subfigure}%
        \hfill
        \begin{subfigure}[align=t]{.325\textwidth}
          \centering
          \includegraphics[width=1\linewidth]{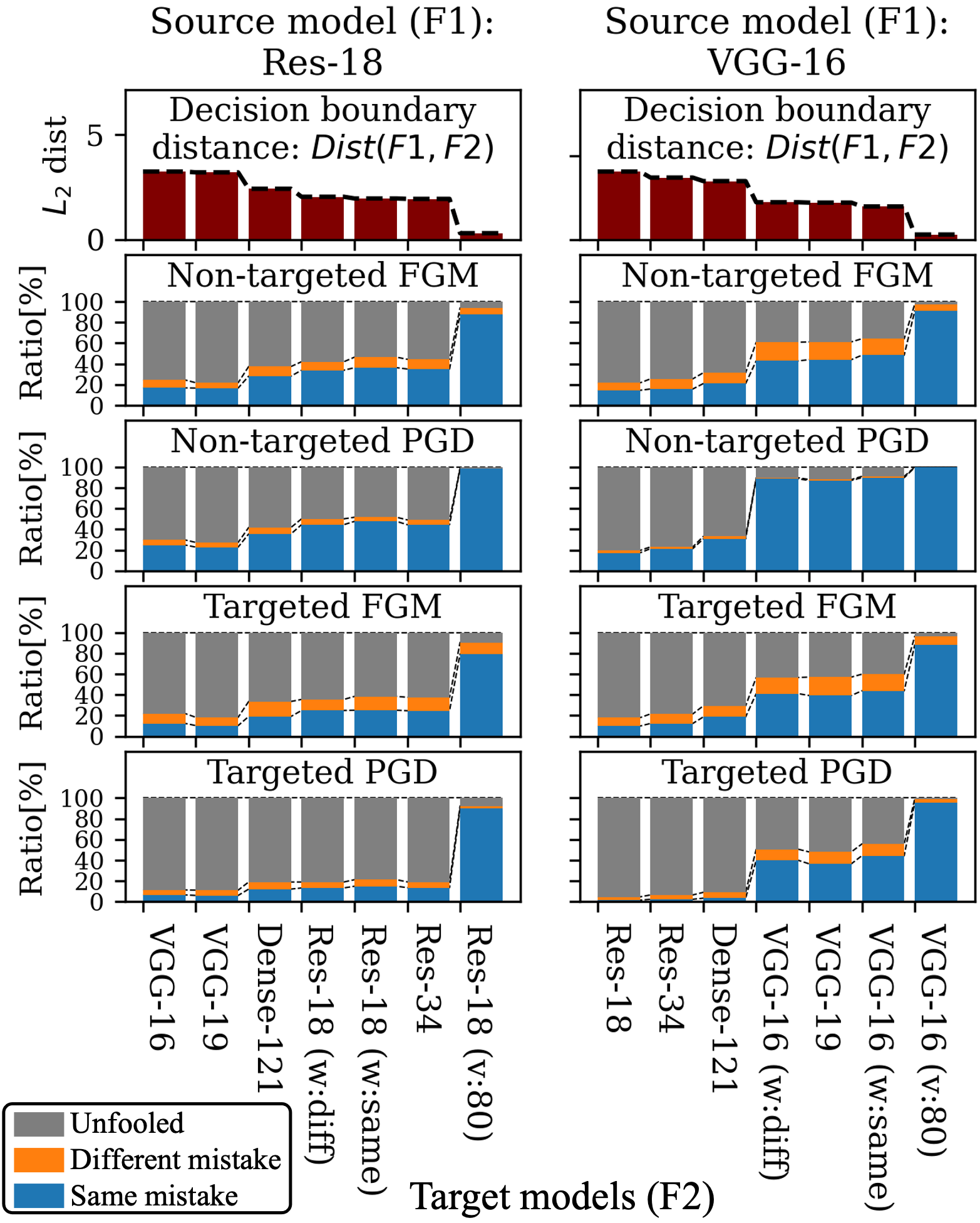}
          \caption[9pt,align=t]{CIFAR-10 ($\epsilon$=1.0)}
          \label{fig:analysis_cifar-10}
        \end{subfigure}%
        \hfill
        \begin{subfigure}[align=t]{.34\textwidth}
          \centering
          \includegraphics[width=1\linewidth]{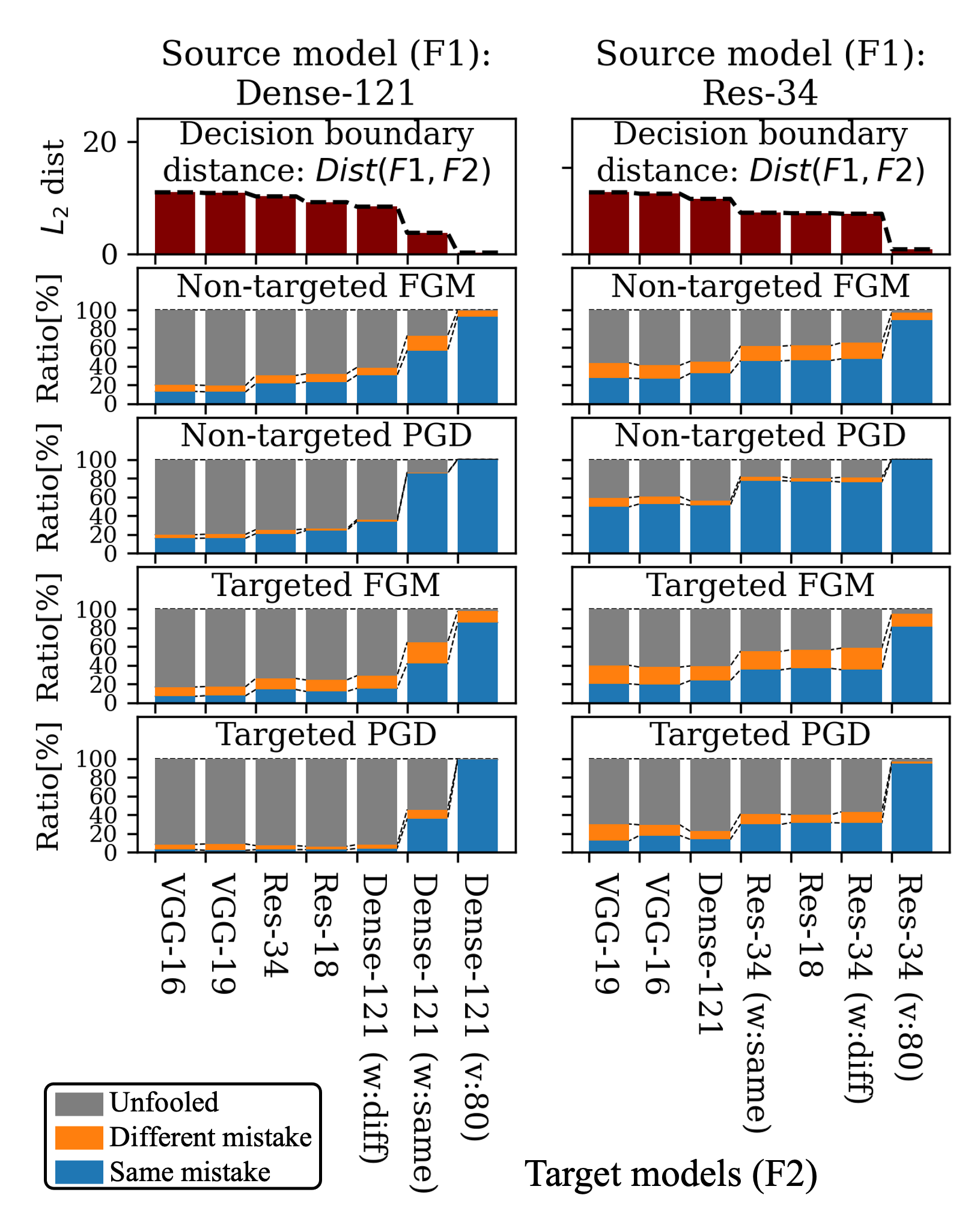}
          \caption[9pt,align=t]{STL-10 ($\epsilon$=5.0)}
          \label{fig:analysis_stl-10}
        \end{subfigure}
  \caption{Class-aware transferability of adversarial attacks against various datasets and models.
  AEs were $l_2$-bounded by the specific $\epsilon$. 
  Order of $F2$ is sorted by $Dist(F1,F2)$ (1st row) for each $F1$ so rightmost $F2$ was estimated to be more similar to F1.
  }
    \label{fig:analysis_summary} 
\end{figure*}

The results of FGM and PGD (ten-step) attacks against various datasets and models with both non-targeted and targeted objectives are shown in Figure~\ref{fig:analysis_summary}.
The $F2$ target models are sorted by quantitative similarity measurement $Dist(F1,F2)$ for each $F1$. 
$Dist(F1,F2)$ roughly corresponds to the qualitative similarity of the models; for example, when $F1$ was ResNet-18, $Dist(F1,F2)$ was the shortest for $F2$ in the ResNet architecture family (Figure~\ref{fig:analysis_cifar-10}).


Figure~\ref{fig:analysis_summary} shows that the majority of the fooled ratio is the same mistake ratio.
Moreover, the same mistake ratio strongly correlates with the fooled ratio:
both fooled and same mistake ratios were higher when the source and target models were in the same architecture family (e.g., ResNet-18 and ResNet-34 are both in the ResNet family) and when the target models were early versions of the source models. 
The correlations between the fooled and same mistake ratios were greater than $0.99$, and the correlations between $Dist(F1, F2)$ and the same mistake ratio were lower than $-0.90$ in all cases shown in Figure~\ref{fig:analysis_summary}. 
It indicates that the fact that AEs tend to cause same mistakes is
strongly connected to their capability to mislead target models' predictions (non-targeted transferability).

\begin{figure}[t]
    \centering
    \includegraphics[width=\linewidth]{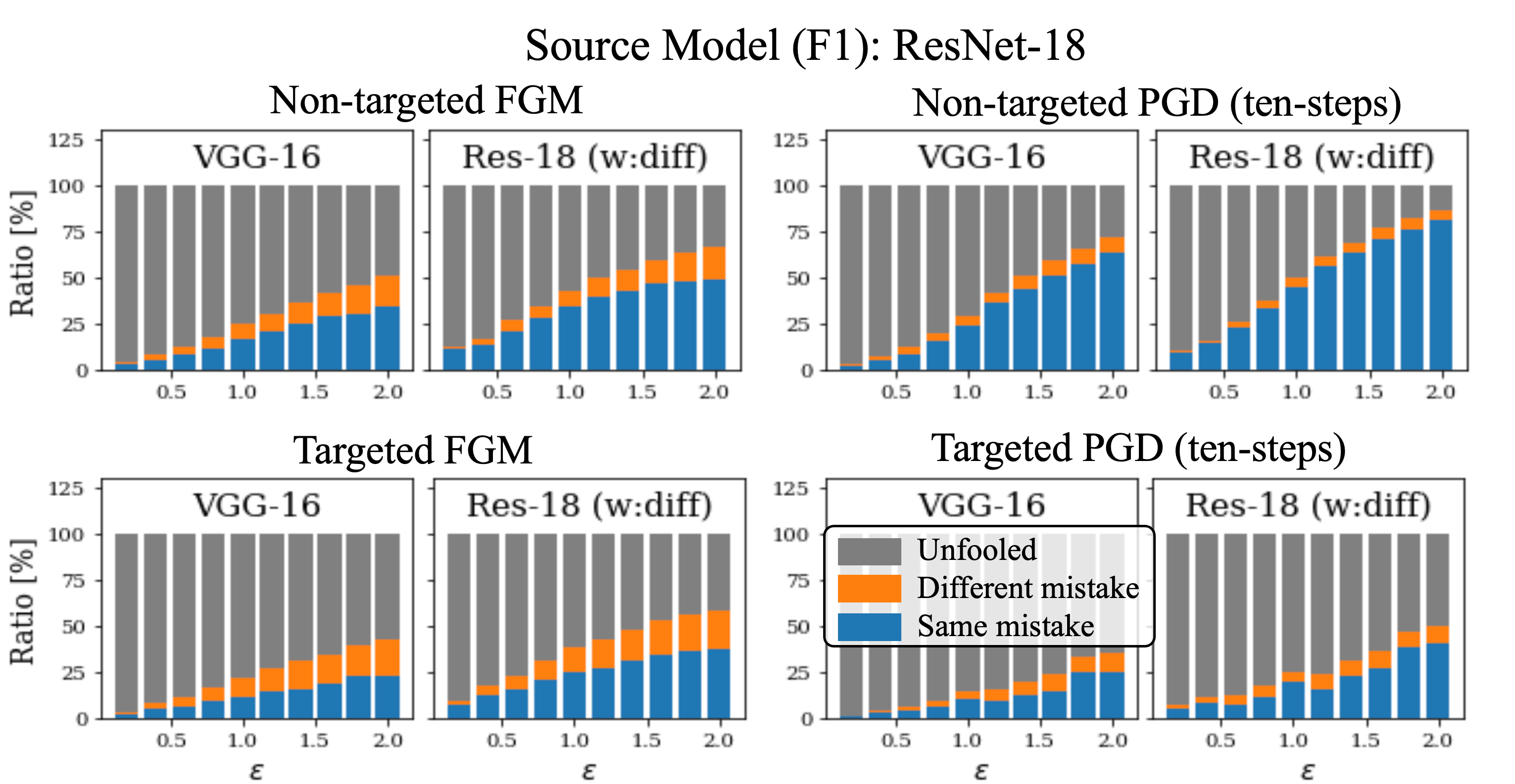}
    \caption{Class-aware transferability of AEs when size of perturbation $\epsilon$ was gradually changed (CIFAR-10).} 
\label{fig:compare_eps} 
\end{figure}

Although AEs tend to cause same mistakes, we observed a non-trivial proportion of different mistakes even when the source and target models were qualitatively very similar (Figure~\ref{fig:analysis_summary}). 
Even when the models had the same architecture and were trained from the same initial weights, the different mistake ratios for targeted FGM attacks were around 20\% for STL-10 (Figure~\ref{fig:analysis_stl-10}). 
Moreover, different mistakes exist even between the source model and the source model at $i^{th}$ epoch.
These findings raise the question of what can explain the presence of different mistakes between similar models, which we address in a later section.

Figure~\ref{fig:compare_eps} shows that, while same mistakes increase with larger perturbations, the different mistake ratio stays almost constant or increases. 
It indicates that there is a misalignment between the ability of AEs to mislead the source model and target model towards a specific class that cannot be resolved simply by enlarging the perturbations. 

\begin{figure}[t]
    \centering
    \includegraphics[width=1\linewidth]{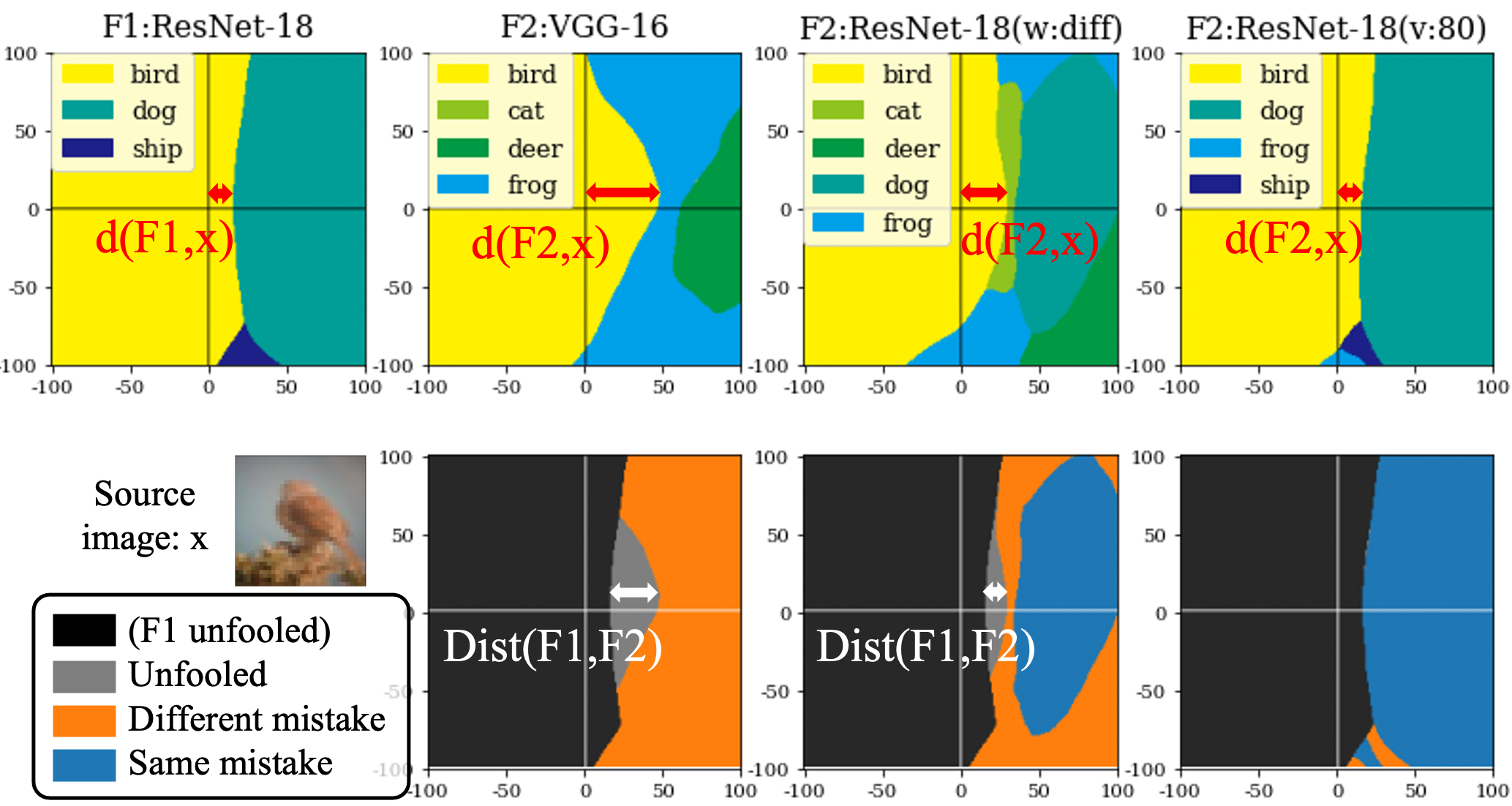}
    \caption{Visualization of decision boundaries for source image of ``bird" in CIFAR-10. 
    First row shows classification results; each color represents a certain class. 
    Second row shows to which areas the three cases of class-aware transferability correspond.
    Distance from (0,0) point to closest decision boundary along x-axis corresponds to metric $d(F1, x)$ described in Section~\ref{subsec:measure_trans}. 
    Unit of each axis is 0.02 in $l_2$ distance.}
\label{fig:decision_boundary} 
\end{figure}

To further interpret these class-aware transferability observations, we visualized the decision boundaries, as in Liu et al. \cite{liu2016delving_into_trans} (Figure~\ref{fig:decision_boundary}). 
We chose two directions of $\delta_1$, the non-targeted gradient direction of ResNet-18, and $\delta_2$, the random orthogonal direction. Both $\delta_1$ and $\delta_2$ were normalized to 0.02 by $l_2$-norm. Each point $(u, v)$ in the 2-D plane corresponded to the image $x+u\delta_1+v\delta_2$, where $x$ is the source image. 
For each model, we plot the classified label of the image corresponding to each point. 
First, we observe an area of different mistakes between the models with the same architecture or even models with only a 20-epoch difference.
Second, the area of same mistakes is larger when the minimum distance to the decision boundary along the x-axis, $d(F_i,x)$, is similar between $F1$ and $F2$. 
It indicates that, while the similarity of the decision boundary separating the true and wrong classes results in non-targeted transferability \cite{liu2016delving_into_trans}, at the same time, the decision boundaries separating different wrong classes can also be similar and can result in same mistakes.

The strong connection between non-targeted transferability and same mistakes indicates the presence of non-robust features \cite{ilyas2019AE_non_rob} in AEs: AEs can cause same mistakes by containing non-robust features that correlate with a specific class. 
However, the presence of different mistakes between similar models or when the perturbations are large is still poorly understood. 
We hypothesized that different mistakes occur when the usage of non-robust features is model-dependent, which we examine in a later section.

\section{Non-robust feature investigation}
\label{sec:non_rob}

\subsection{Overview}

Here, we provide the first possible explanation for different mistakes, one that can also explain same mistakes. 
Specifically, we provide insightful explanations and discussions based on the theory of non-robust features \cite{ilyas2019AE_non_rob}. 
Same mistakes can be due to different models using similar non-robust features; we show that a different mistake can also arise from non-robust features. 

We designed N-targeted attack to generate AEs that can cause different mistakes for different models. 
Then by using Ilyas et al.'s framework \cite{ilyas2019AE_non_rob}, we show that those AEs have non-robust features of two different classes to which those models were misled. 
Our results indicate that two models can make different mistakes when they use the non-robust features of those two classes differently.
We thus conclude that the usage of non-robust features is a possible explanation for different and same mistakes: Same mistakes are due to AEs having the non-robust features of the class to which a model was misled; on the other hand, AEs may simultaneously have multiple non-robust features that correlate with different classes, and if models use them differently, they may classify the same AEs differently (Figure~\ref{fig:hypothesis}). 

\begin{figure*}[t]
    \centering
    \includegraphics[width=0.80\linewidth]{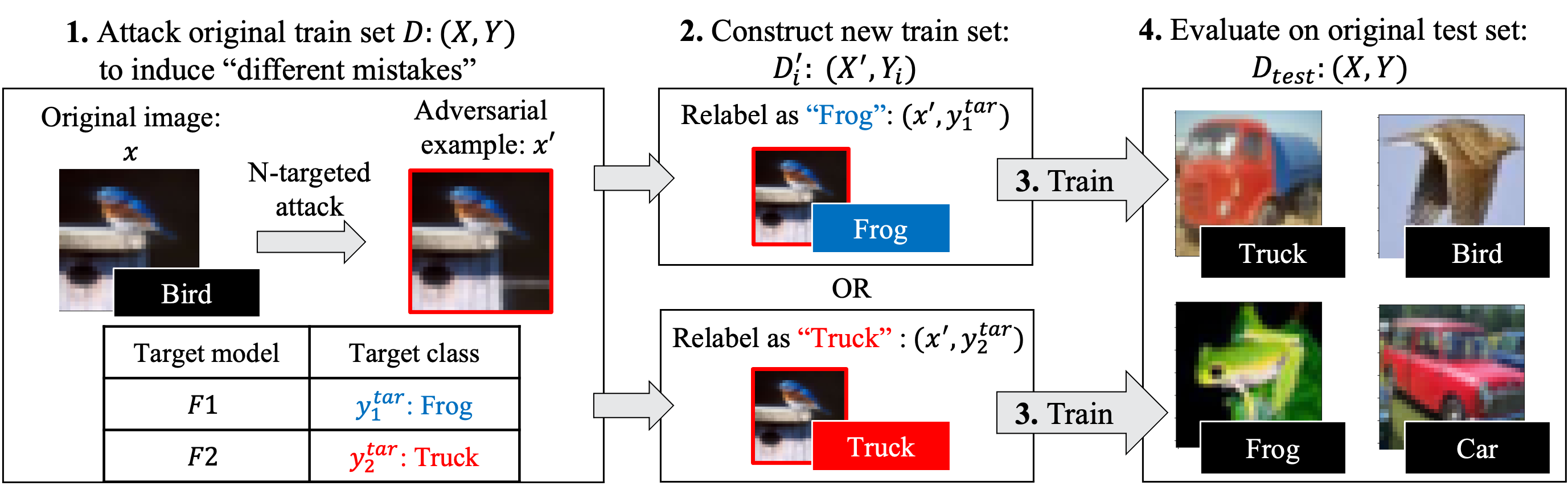} 
    \caption{Illustration of experiment to test our hypothesis (Figure~\ref{fig:hypothesis}) that different mistakes can arise from AEs having non-robust features of two different classes to which two different models were misled. 
    First, original training set is attacked by N-targeted attack to generate AEs that induce different mistakes for $F1$ or $F2$. 
    Next, a new (non-robust) dataset is constructed by relabeling generated AEs as either $Y1$ or $Y2$, the target classes for $F1$ or $F2$. 
    Finally, models are trained on new datasets and evaluated on original test set.}
    \label{fig:illust_non_rob_exp}
\end{figure*}

\subsubsection{Experiment.} 
We aim to detect non-robust features of two different classes from the AEs that caused different mistakes. 
Suppose those AEs have non-robust features of two misleading classes simultaneously. 
In that case, we can assume that the models used the non-robust features differently (as in Figure~\ref{fig:hypothesis}). 
To demonstrate the presence of non-robust features in AEs, we used the framework for non-robust features described by Ilyas et al. \cite{ilyas2019AE_non_rob}. 
We aim to detect non-robust features of two different classes from AEs, whereas Ilyas et al. \cite{ilyas2019AE_non_rob} did it with only one class.

The flow of the experiment is illustrated in Figure~\ref{fig:illust_non_rob_exp}. 
First, we generate AEs that can cause different mistakes for models $F1$ and $F2$, on the original training set: each AE $x'$ is generated from the original image $x$ to mislead a model $F1$ to a target class $y^{tar}_1$ and a model $F2$ to a target class $y^{tar}_2$. 
Then we created new (non-robust) training sets using two ways of relabeling the whole set of AEs $X'$, i.e., by either of the corresponding target classes $Y1$ or $Y2$ (note that $X$, $X'$, $Y1$, and $Y2$ are the collections of the datapoints $x$, $x'$, $y^{tar}_1$, and $y^{tar}_2$, respectively). 
Here, the target classes $Y1$ and $Y2$ are randomly chosen for each data point so that only the non-robust features of specific classes may correlate with the assigned labels, but other features have approximately zero correlation, as in Ilyas et al. \cite{ilyas2019AE_non_rob}.
Finally, we trained a model on the new training set ($D'_1: (X', Y1)$ or $D'_2: (X', Y2)$) and evaluated it on the original test set, $D_{test}: (X, Y)$. 
If both non-robust sets $D'_1$ and $D'_2$ were useful in generalizing to the original test set $D_{test}$, we can conclude that non-robust features of both classes ($Y1$ and $Y2$) are present in the same AEs at the same time.

\begin{figure}[h]
    \centering
        \includegraphics[width=1.0\linewidth]{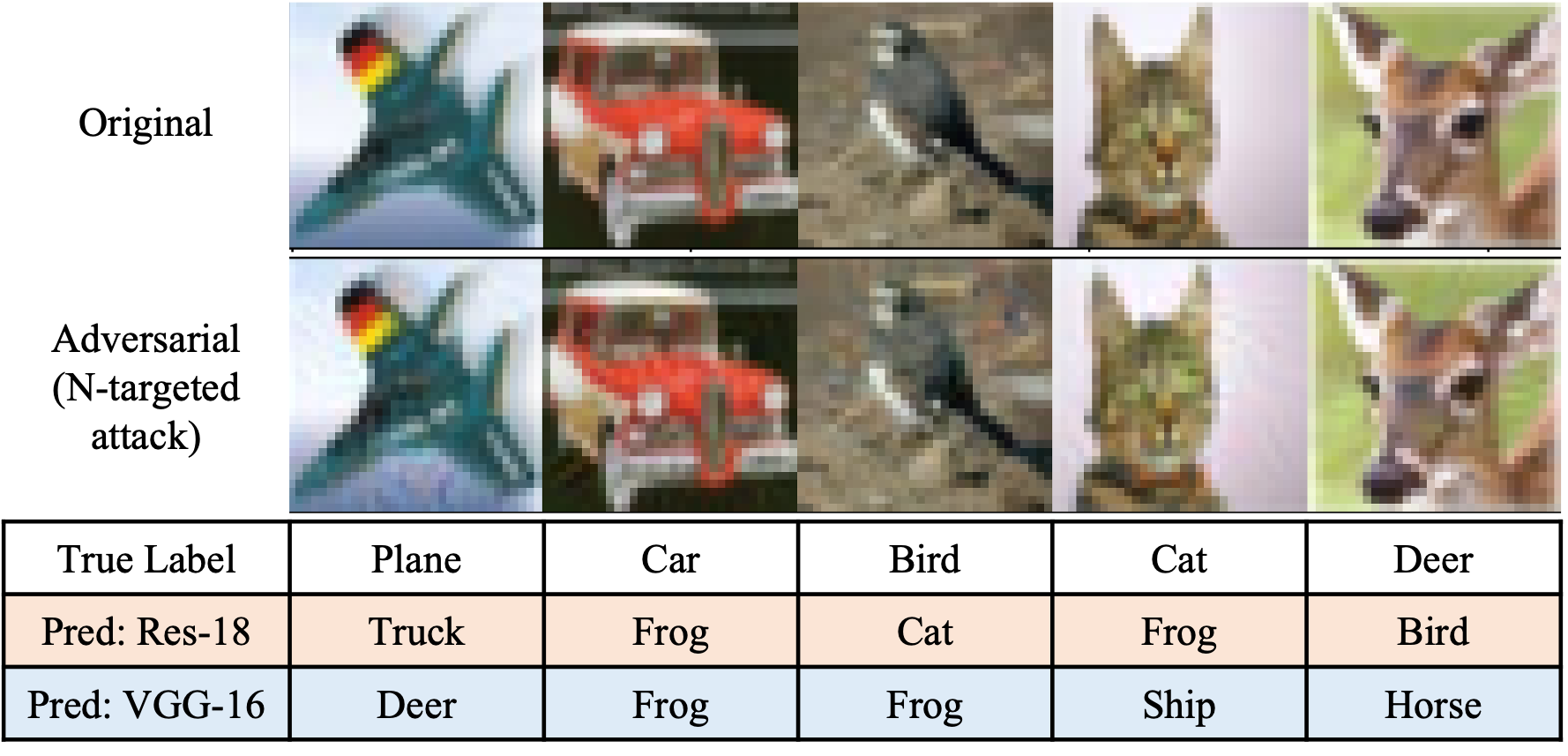} 
        \caption{Examples of AEs (lower row) generated for images from CIFAR-10 (upper row) generated by N-targeted attack that was $l_2$-bounded by $\epsilon$=1.0.
        ResNet-18 and VGG-16 correctly classified the original images, whereas the AEs misled the models toward two random classes (For the entire set, over 90\% were successful).
        The entire set of generated AEs comprises a non-robust set by relabeling them, as illustrated in Figure~\ref{fig:illust_non_rob_exp}.}
        \label{fig:n_tar_examples}
\end{figure}

We generated AEs that can cause different mistakes for $F1$ and $F2$ by using our extended version of a targeted attack, namely \textit{N-targeted attack}. 
This attack is aimed at misleading model $F_i$ towards each target class $y^{tar}_i$.
The objective of an N-targeted attack is represented as
\small
\begin{equation}
    \label{eq:n_tar}
    \centering
     \argmin_{x'} \sum_{i=1}^N L \left( F_i(x'), y_i^{tar} \right), \;\; \text{s.t.} \;\; \| x' - x \|_p < \epsilon.
    \end{equation}
\normalsize
It simply sums up all the loss values for all target models. 
The optimization problem is solved iteratively using the same algorithm as PGD. 
The generated AEs for \{F1, F2\}=\{ResNet-18, VGG-16\} are shown in Figure~\ref{fig:n_tar_examples}.

\subsection{Experiment Settings}

\subsubsection{Non-robust Set}
We construct non-robust sets for Fashion-MNIST, CIFAR-10, and STL-10, using the models used in Section \ref{sec:analysis}.
Non-robust training sets were constructed using an N-targeted attack based on PGD-based optimization with 100 steps with step size $\alpha$=0.1 (For STL-10, we generated ten AEs per image to increase the data size from 5,000 to 50,000).
AEs were $l_2$-bounded by $\epsilon$ of 2.0, 1.0, and 5.0 for Fashion-MNIST, CIFAR-10, and STL-10.

Note that AEs generated by N-targeted attack could simultaneously lead predictions of models F1 and F2 towards different classes Y1 and Y2 at a high rate: 60\% for Fashion-MNIST and over 90\% for CIFAR-10 and STL-10.
It means that it is easy in a white-box setting to generate AEs that cause different predictions for different models, which is particularly interesting.

\subsubsection{Training Models on Non-robust Set}
The optimizer was SGD with momentum set to $0.9$ and weight decay set to $0.0005$, with learning rate decay. 
The initial learning rate, batch size, and data augmentation were optimized using a grid search. 
We trained FC-2 and Conv-2 (described in Section~\ref{sec:analysis}) for Fashion-MNIST, and ResNet-18 and VGG-16\_bn (VGG-16 with batch normalization) for CIFAR-10 and STL-10.

\subsection{Results and Discussions}

\begin{table}[t!]
\footnotesize
\centering
    \begin{tabular}{c|l|c|c|c}
        \toprule
            Dataset & \begin{tabular}[c]{@{}c@{}}Non-robust set \\ constructed for\end{tabular} & \begin{tabular}[c]{@{}c@{}}Train\\ set\end{tabular} & \begin{tabular}[c]{@{}c@{}}Trained\\ model\end{tabular} & \begin{tabular}[c]{@{}c@{}}Test acc\\(X, Y)\end{tabular} \\ 
        \midrule
            \multirow{4}{*}{\begin{tabular}[c]{@{}c@{}}Fashion-\\ MNIST\end{tabular}} & \multirow{4}{*}{\begin{tabular}[c]{@{}l@{}}F1: Conv-2\\ F2: FC-2\end{tabular}} & \multirow{2}{*}{$D'_1$: (X', Y1)} & Conv-2 & 82.9 \\ \
             &  &  & FC-2 & 62.5 \\ \cline{3-5} 
             &  & \multirow{2}{*}{$D'_2$: (X', Y2)} & Conv-2 & 80.3 \\
             &  &  & FC-2 & 75.4 \\ \hline
            \multirow{8}{*}{\begin{tabular}[c]{@{}c@{}}CIFAR-\\ 10\end{tabular}} & \multirow{4}{*}{\begin{tabular}[c]{@{}l@{}}F1: Res-18\\ F2: VGG-16\end{tabular}} & \multirow{2}{*}{$D'_1$: (X', Y1)} & Res-18 & 51.3 \\ 
             &  &  & VGG-16\_bn & 53.9 \\ \cline{3-5} 
             &  & \multirow{2}{*}{$D'_2$: (X', Y2)} & Res-18 & 10.2 \\
             &  &  & VGG-16\_bn & 71.0 \\ \cline{2-5} 
             & \multirow{4}{*}{\begin{tabular}[c]{@{}l@{}}F1: Res-18\\ F2: Res-18\\ \qquad(w:same)\end{tabular}} & \multirow{2}{*}{$D'_1$: (X', Y1)} & Res-18 & 50.1 \\ 
             &  &  & VGG-16\_bn & 54.1 \\ \cline{3-5} 
             &  & \multirow{2}{*}{$D'_2$: (X', Y2)} & Res-18 & 59.2 \\ 
             &  &  & VGG-16\_bn & 58.9 \\ \hline
            \multirow{8}{*}{\begin{tabular}[c]{@{}c@{}}STL-\\ 10\end{tabular}} & \multirow{4}{*}{\begin{tabular}[c]{@{}l@{}}F1: Res-18\\ F2: VGG-16\end{tabular}} & \multirow{2}{*}{$D'_1$: (X', Y1)} & Res-18 & 24.0 \\
             &  &  & VGG-16\_bn & 25.4 \\ \cline{3-5} 
             &  & \multirow{2}{*}{$D'_2$: (X', Y2)} & Res-18 & 53.7 \\
             &  &  & VGG-16\_bn & 56.0 \\ \cline{2-5} 
             & \multirow{4}{*}{\begin{tabular}[c]{@{}l@{}}F1: VGG-16\\ F2: VGG-16\\ \qquad(w:same)\end{tabular}} & \multirow{2}{*}{$D'_1$: (X', Y1)} & Res-18 &  38.4\\ 
             &  &  & VGG-16\_bn & 51.8 \\ \cline{3-5} 
             &  & \multirow{2}{*}{$D'_2$: (X', Y2)} & Res-18 & 52.2 \\ 
             &  &  & VGG-16\_bn & 52.4 \\ 
        \bottomrule
    \end{tabular}
    \caption{Test accuracy on original test set when model was trained on non-robust sets. 
    Non-robust set $D'_i$ contains AEs generated by N-targeted attack and relabeled as $Yi$, the target classes for model $Fi$. 
    Since the random accuracy of 10-class dataset is 10\%, we can say that models were generalized to original test set by training on non-robust sets. 
    It shows that the generated AEs that induced different mistakes at a high rate contain multiple non-robust features that correlate with two different classes simultaneously.
    }
    \label{table:train_non_rob}
\end{table}

Table~\ref{table:train_non_rob} shows the test accuracies of the models trained on the constructed non-robust sets. 
For all pairs of attacked models $F1$ and $F2$, the test accuracies on the original test set $(X, Y)$ were higher than the random accuracy of 10\% for both relabeling cases ($Y1$ or $Y2$). 
This result shows that the models could learn non-robust features of $Y1$ by training on the non-robust set $D'_1: (X', Y1)$ and non-robust features of $Y2$ by training on the non-robust set $D'_2: (X', Y2)$. 
In other words, it is shown that the generated AEs $X'$ had multiple non-robust features that correlate with two different classes simultaneously. 
This result supports our hypothesis: different mistakes can arise when AEs have non-robust features of two classes and when models use them differently. 
It is interpreted that the ratio of different mistakes did not decrease with larger perturbations (Figure~\ref{fig:compare_eps}) because it did not resolve the misalignment of non-robust feature usage between models.

In addition, Table~\ref{table:train_non_rob} shows that even models with the same architecture and initial weight parameters learn non-robust features differently.
It suggests that learned non-robust features differ only with the stochasticity of updating the weight parameters caused by a shuffled training set or dropout layers, which explains the presence of different mistakes between models with high similarity in Figure~\ref{fig:analysis_summary}.

\section{Transferability of AEs generated for ensemble models}
\label{sec: ensemble}
Section~\ref{sec:non_rob} indicates that different mistakes can occur when models use non-robust features differently. 
Therefore, different mistakes are expected to decrease when AEs contain ``general" non-robust features used by many models.

To verify this, we generate targeted AEs for an ensemble of models: those AEs should contain only non-robust features that are ``agreed" to be correlated with the target classes by different models.
Liu et al. \cite{liu2016delving_into_trans} showed that attacking an ensemble model can improve targeted transferability; we reveal that non-robust features can explain it.
In Table~\ref{tab:ensemble}, we compare targeted AEs generated for a single model (i.e., Vanilla attack) and an ensemble model (i.e., Ensemble attack) using PGD.
The source model F1 is ResNet-18, and the target model F2 is VGG-16.
We confirmed that different mistakes decrease when Densenet-121 is additionally used in the Ensemble attack, while same mistakes increase.
In contrast, the Ensemble attack increased the number of AEs that did not fool F1 (F1 unfooled): since the Ensemble attack tries to inject only non-robust features commonly used by models, it sacrifices the use of model-specific non-robust features used by F1.

\begin{table}[ht]
\footnotesize
\centering
    \begin{tabular}{c|cccc}
        \toprule
            \multirow{2}{*}{Attack} & \multirow{2}{*}{\begin{tabular}[c]{@{}c@{}}F1: \\ unfooled\end{tabular}} & \multicolumn{3}{c}{F1: fooled} \\ \cline{3-5} 
            &  & \begin{tabular}[c]{@{}c@{}}F2:\\ unfooled\end{tabular} & \begin{tabular}[c]{@{}c@{}}F2: different\\ mistake\end{tabular} & \begin{tabular}[c]{@{}c@{}}F2: same\\ mistake\end{tabular} \\
        \midrule
            Vanilla & 414 & 3910 & 250 & 426 \\ \hline
            \begin{tabular}[c]{@{}c@{}}Ensemble\\ (+ Dense-121)\end{tabular} & 1988 & 1954 &  216 (-34) &  842 (+416)\\
        \bottomrule
    \end{tabular}
    \caption{Comparison between Vanilla and Ensemble targeted attack on CIFAR-10.
    AEs were $l_2$-bounded by $\epsilon$=1.0, generated by targeted PGD (ten-step). 
    Source model F1 is ResNet-18, and target model F2 is VGG-16.
    Each value shows the number of each case for randomly selected 5,000 images. 
    For a fair comparison, we used the same target class for each image for both attacks.}
    \label{tab:ensemble}
\end{table}

\section{Conclusion}
\label{sec:conclusion}


We demonstrated that AEs tend to cause same mistakes, which is consistent with the fact that AEs can have non-robust features that correlate with a certain class.
However, we further showed that different mistakes could occur between similar models regardless of the perturbation size, raising the question of how AEs cause different mistakes. 

We indicate that non-robust features can explain both different and same mistakes. 
Ilyas et al. \cite{ilyas2019AE_non_rob} showed that AEs can have non-robust features that are predictive but are human-imperceptible, which can cause same mistakes. 
In contrast, we reveal a novel insight that different mistakes occur when models use non-robust features differently.

Future work includes developing transferable adversarial attacks based on our findings: AEs should transfer when they contain non-robust features commonly used by different DNNs.
In addition, since we do not conclude that all same mistakes and different mistakes are due to non-robust features, whether there is another mechanism is an important research question.

\hfill \break
\noindent
\textbf{Acknowledgements: }
This work was partially supported by JSPS KAKENHI Grants JP16H06302, JP18H04120, JP20K23355, JP21H04907, and JP21K18023, and by JST CREST Grants JPMJCR18A6 and JPMJCR20D3, Japan.

\clearpage

{\small
\bibliographystyle{ieee_fullname}
\bibliography{egpaper}
}

\clearpage


\appendix

\section{Details of datasets}
In Table~\ref{tab:dataset_detail}, we provide the details of the datasets we used in the main paper.

\begin{table}[!htbp]
\centering
\footnotesize
    \begin{tabular}{c|c|c|c|c}
        \toprule
        Dataset & Class num. & Image size & Train & Test \\
        \midrule
        Fashion-MNIST & 10 & (1,28,28) & 60,000 & 10,000 \\
        CIFAR-10 & 10 & (3,32,32) & 50,000 & 10,000 \\
        STL-10 & 10 & (3,96,96) & 5,000 & 8,000 \\
        \bottomrule
    \end{tabular}
\caption{Details of datasets we used in the main paper.
    Image size represents (channel, height, width) of images.
}
\label{tab:dataset_detail}
\end{table}

\section{Adversarial Transferability Analysis}

In this section, we provide supplementary results for our analysis of the class-aware transferability of AEs. 

\subsection{Details of evaluated models}
All models were trained using the stochastic gradient descent (SGD) optimizer with a momentum of $0.9$ and weight decay of $0.0005$. 

For Fashion-MNIST, we trained models at an initial learning rate of $0.01$, which decayed $0.1$ times at the $20^{th}$ epoch with 40 epochs in total.
Details of model architectures of FC-2/-4 and Conv-2/-4 are described in Table~\ref{tab:detail_arch_fashion_mnist}.

For CIFAR-10 and STL-10, we trained models at an initial learning rate of $0.01$, which decayed $0.1$ times at the $50^{th}$ epoch with 100
epochs in total.
Additionally, we used data augmentation techniques to train models for CIFAR-10 and STL-10 to prevent strong overfit.

\begin{table*}[p]
\centering
\footnotesize
\begin{tabular}{c|c|c|c}
\toprule
FC-2 & FC-4 & Conv-2 & Conv-4 \\
\midrule
\begin{tabular}[c]{@{}c@{}}Linear: (784, 500)\\ ReLU\\ Linear: (500, 10)\end{tabular} & \begin{tabular}[c]{@{}c@{}}Linear: (784, 500)\\ ReLU\\ Linear: (500, 200)\\ ReLU\\ Linear: (200, 100)\\ ReLU\\ Linear: (100, 10)\end{tabular} & \begin{tabular}[c]{@{}c@{}}Conv2d: (1, 32, 3, 1)\\ ReLU\\ Conv2d: (32, 64, 3, 1)\\ ReLU\\ Maxpool\\ Linear: (9216, 128)\\ ReLU\\ Dropout\\ Linear: (128, 10)\end{tabular} & \begin{tabular}[c]{@{}c@{}}Conv2d: (1, 32, 3, 1) \\ ReLU\\ Conv2d: (32, 64, 3, 1)\\ ReLU\\ Conv2d: (64, 128, 3, 1)\\ ReLU\\ Maxpool\\ Conv2d: (128, 128, 3, 1)\\ ReLU\\ Maxpool\\ Linear: (9216, 128)\\ ReLU\\ Dropout\\ Linear: (128, 10)\end{tabular} \\
\bottomrule
\end{tabular}
\caption{Model architectures used for Fashion-MNIST.
    ``Linear: $(i, j)$" is a fully connected layer with input size $i$ and output size $j$. 
    ``Conv2d: $(C_i, C_o, k, s)$" is a convolution layer with input channel size $C_i$, output channel size $C_o$, kernel size $k$, and stride $s$.
}
\label{tab:detail_arch_fashion_mnist}
\end{table*}

\subsection{Model Similarity Analysis}
\begin{figure*}[p]
    \centering
        \includegraphics[width=0.7\linewidth]{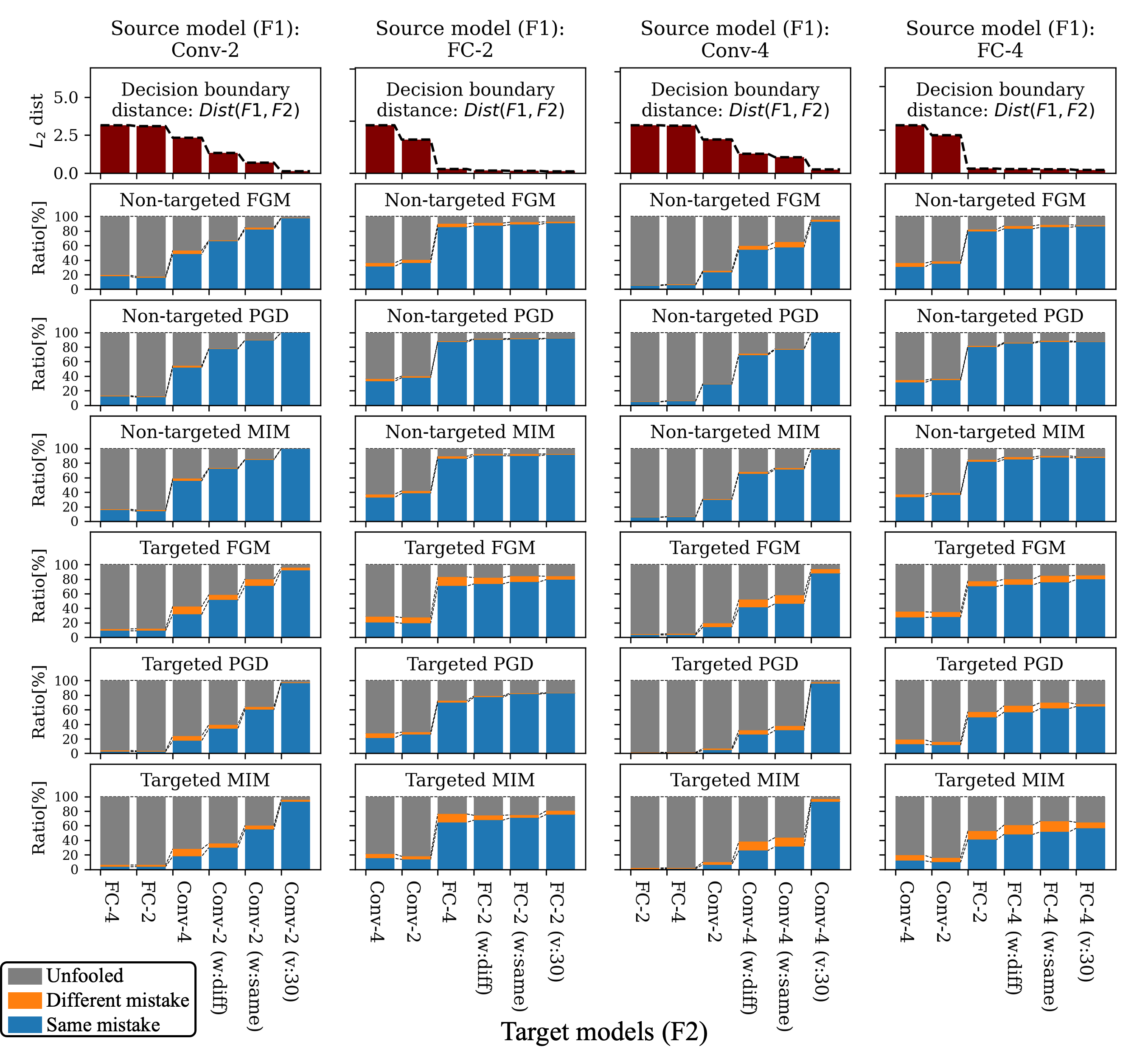} 
        \caption{Class-aware transferability of adversarial attacks for Fashion-MNIST. 
        We evaluate FGM \cite{goodfellow2014explaining_harnessing_ae}, PGD \cite{madry2017towards_pgd_at_adversarial_training}, and MIM \cite{dong2018boosting_trans_momentum} with both non-targeted and targeted objectives.
        AEs were l2-bounded by $\epsilon$=1.0. 
        Order of F2 is sorted by $Dist(F1, F2)$ (1st row) for each F1 so rightmost F2 was estimated to be more similar to F1.}
        \label{fig:analysis_f_mnist_all}
\end{figure*}

\begin{figure*}[p]
    \centering
        \includegraphics[width=\linewidth]{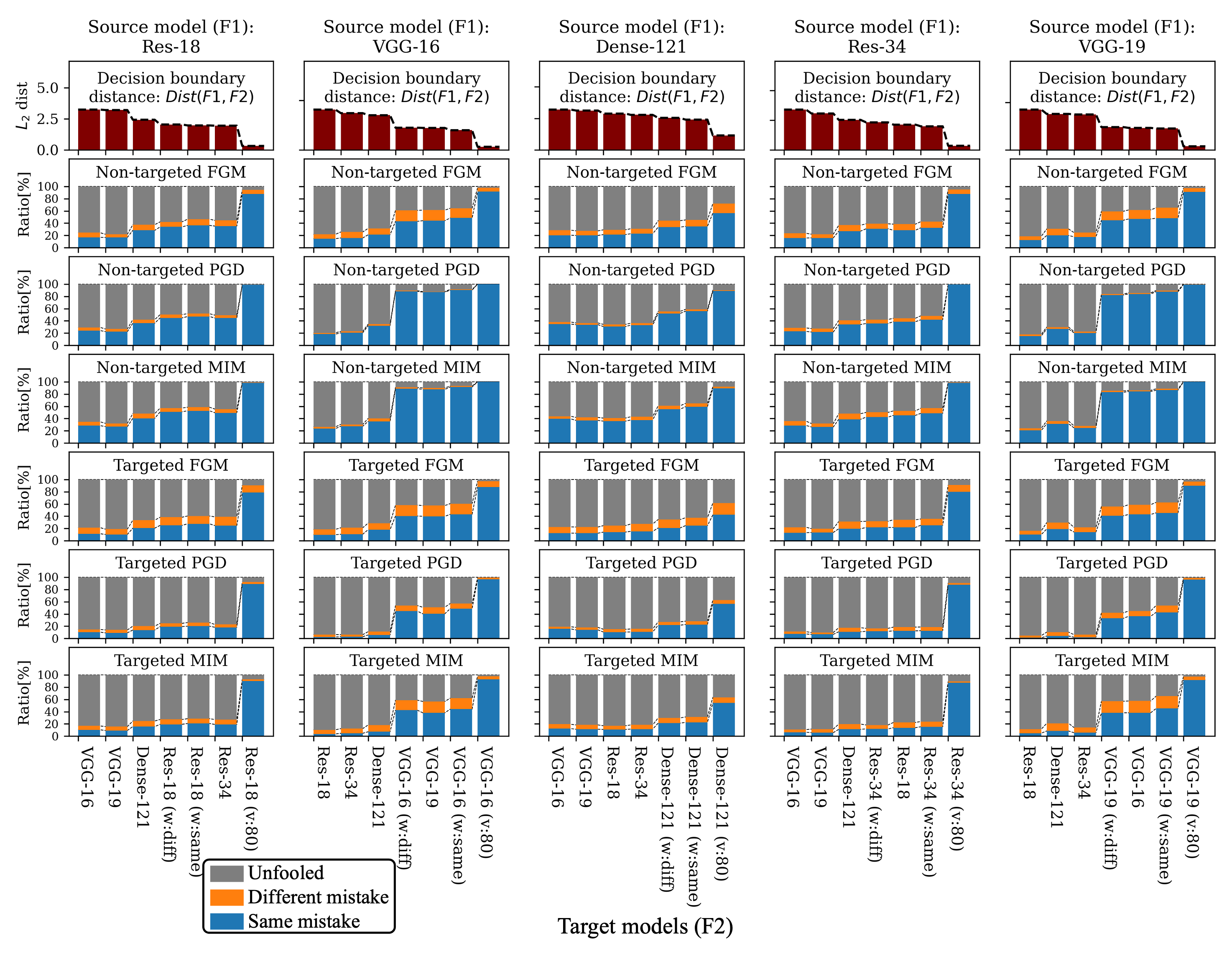} 
        \caption{Class-aware transferability of adversarial attacks for CIFAR-10. 
        We evaluate FGM \cite{goodfellow2014explaining_harnessing_ae}, PGD \cite{madry2017towards_pgd_at_adversarial_training}, and MIM \cite{dong2018boosting_trans_momentum} with both non-targeted and targeted objectives.
        AEs were l2-bounded by $\epsilon$=1.0. 
        Order of F2 is sorted by $Dist(F1, F2)$ (1st row) for each F1 so rightmost F2 was estimated to be more similar to F1..}
        \label{fig:analysis_cifar_all}
\end{figure*}

\begin{figure*}[p]
    \centering
        \includegraphics[width=\linewidth]{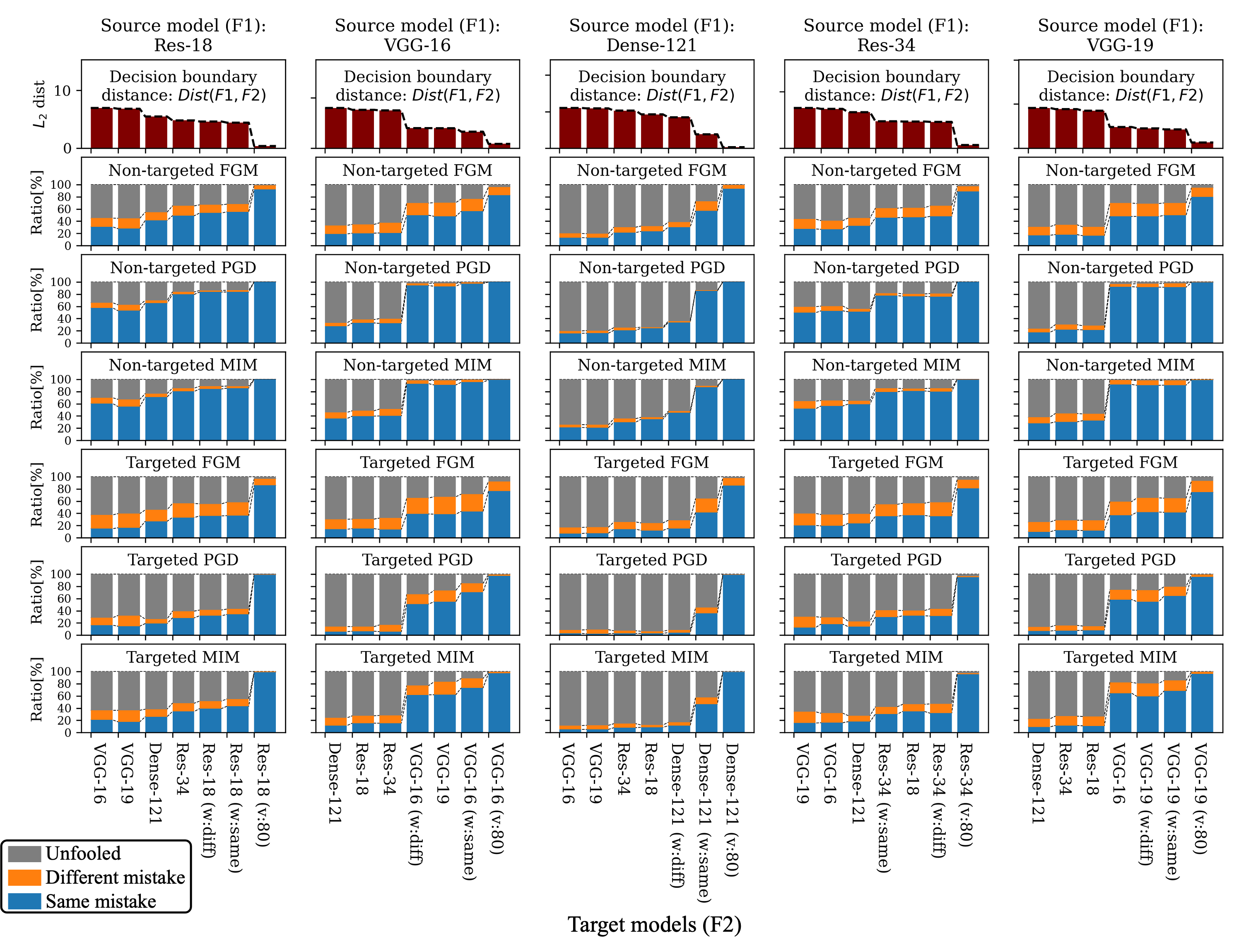} 
        \caption{Class-aware transferability of adversarial attacks for STL-10. 
        We evaluate FGM \cite{goodfellow2014explaining_harnessing_ae}, PGD \cite{madry2017towards_pgd_at_adversarial_training}, and MIM \cite{dong2018boosting_trans_momentum} with both non-targeted and targeted objectives.
        AEs were l2-bounded by $\epsilon$=5.0. 
        Order of F2 is sorted by $Dist(F1, F2)$ (1st row) for each F1 so rightmost F2 was estimated to be more similar to F1.}.
        \label{fig:analysis_stl_all}
\end{figure*}

Here, we show the supplementary results of class-aware transferability of AEs.
The results for Fashion-MNIST, CIFAR-10, and STL-10 are shown in 
Figure~\ref{fig:analysis_f_mnist_all}, Figure~\ref{fig:analysis_cifar_all}, and Figure~\ref{fig:analysis_stl_all}, respectively.
These results include the analysis of various models, which are not in the main paper.  
Furthermore, the analysis of the AEs generated by Momentum Iterative Method (MIM) \cite{dong2018boosting_trans_momentum} is added.
We confirm the consistency of our findings for various models and attacks: the fact that AEs tend to cause same mistakes, but a non-trivial proportion of different mistakes exist is consistent. 

\begin{figure*}[p]
    \centering
        \includegraphics[width=\linewidth]{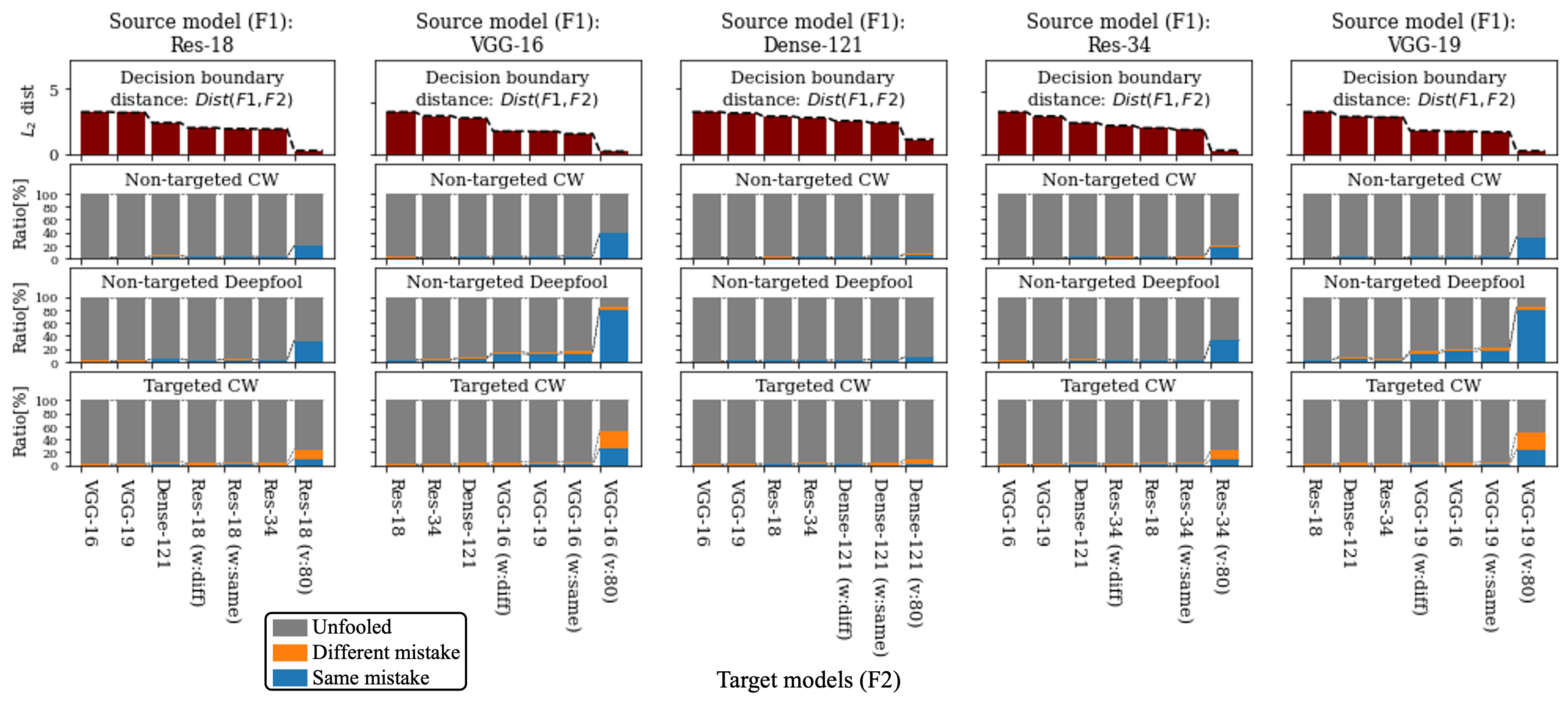} 
        \caption{Class-aware transferability of optimization-based adversarial attacks for CIFAR-10. 
        We evaluate CW \cite{carlini2017towards_cw_attack} and Deepfool \cite{moosavi2016deepfool} (Deepfool is defined only for a non-targeted objective).
        Order of F2 is sorted by $Dist(F1, F2)$ (1st row) for each F1 so rightmost F2 was estimated to be more similar to F1.
        Since these optimization-based attacks try to find minimum perturbations that are enough to fool the source model F1, they hardly transfer between models.
        }.
        \label{fig:analysis_cifar_optim}
\end{figure*}

We also evaluated two optimization-based adversarial attacks, CW \cite{carlini2017towards_cw_attack} and Deepfool \cite{moosavi2016deepfool}.
Figure~\ref{fig:analysis_cifar_optim} shows the results for CIFAR-10.
Since these optimization-based attacks try to find minimum perturbations that are enough to fool the source model F1, they hardly transferred between models.
Interestingly, AEs generated by the optimization-based attacks do not transfer even to the source models at $80^{th}$ epochs.
Therefore, fooled ratios were too small to analyze the class-aware transferability of the AEs.

\subsection{Correlation Between Decision Boundaries' Distance and Class-aware Transferability}
\begin{figure*}[p]
    \centering
        \begin{subfigure}[b]{0.6\linewidth}
            \centering
            \includegraphics[width=\linewidth]{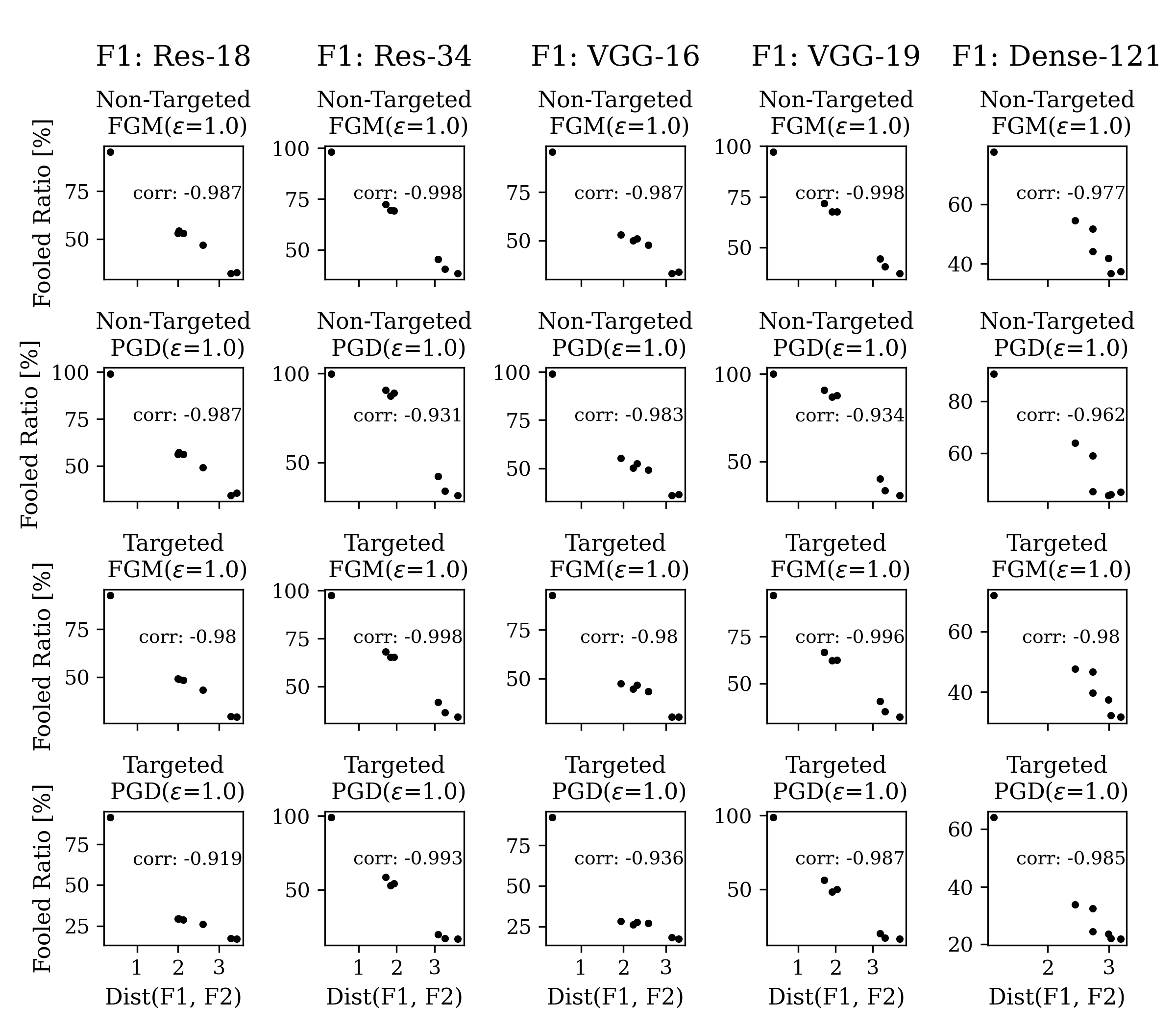} 
            \caption{Correlation between $Dist(F1, F2)$ and fooled ratio for CIFAR-10.
            }
            \label{fig:corr_dist_fooled}
        \end{subfigure}
    \quad
        \begin{subfigure}[b]{0.6\linewidth}
            \centering
            \includegraphics[width=\linewidth]{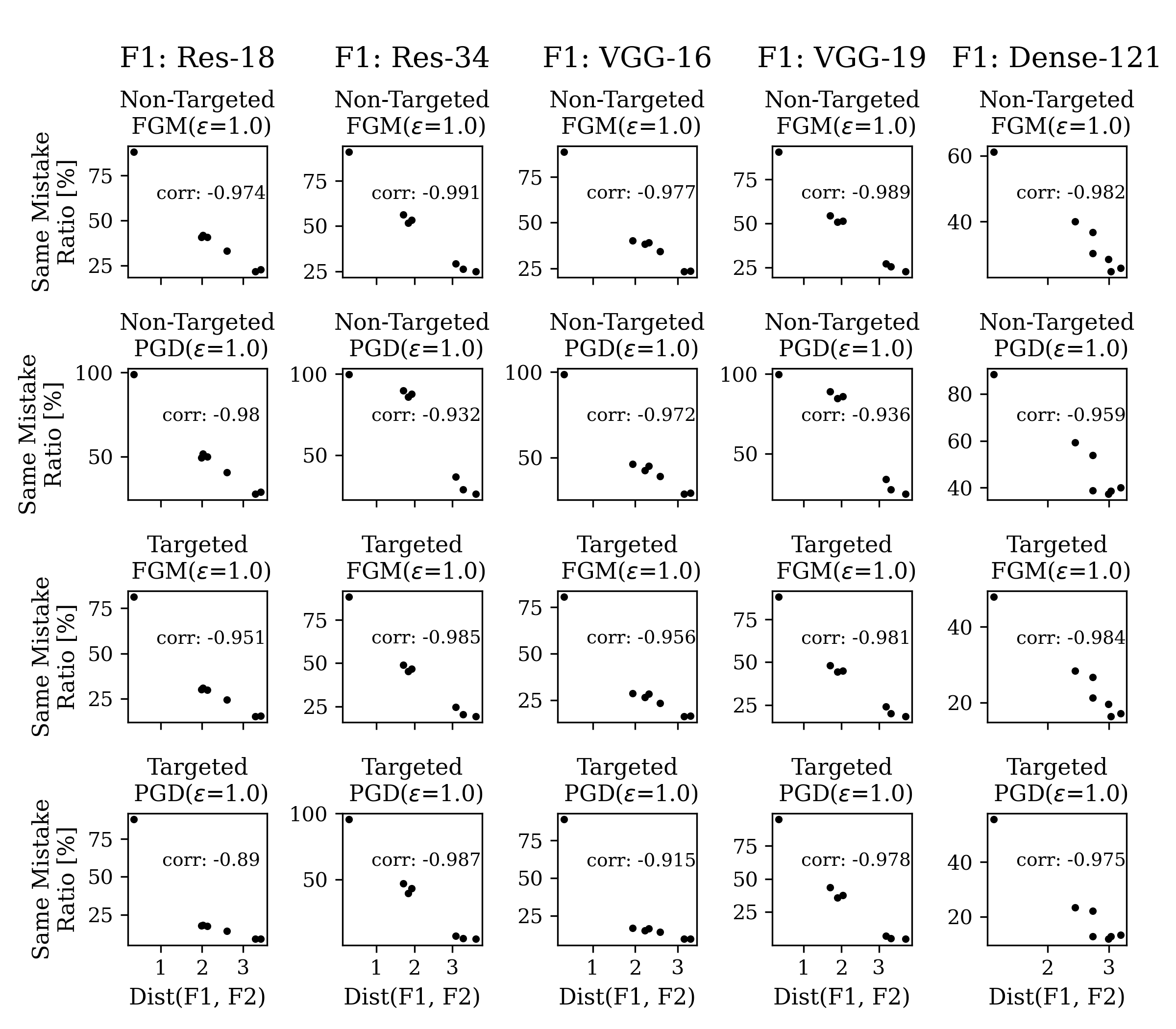}
            \caption{Correlation between $Dist(F1, F2)$ and same mistake ratio for CIFAR-10.
            }
            \label{fig:corr_dist_same}
        \end{subfigure}
        
    \caption{Correlations (1) between $Dist(F1, F2)$ and fooled ratio ratio (Figure~\ref{fig:corr_dist_fooled}), and (2) between $Dist(F1, F2)$ and same mistake (Figure~\ref{fig:corr_dist_same}) for CIFAR-10.
    ResNet-18, ResNet-34, VGG-16, VGG-19, and DenseNet-121 source models (1st to 5th columns, respectively) were attacked by non-targeted and targeted attack (1-2nd and 3-4th rows, respectively) using FGM and PGD (ten-step) (1,3rd and 2,4th rows, respectively) methods. 
    For each source model, the results of the corresponding seven target models (shown in Figure~\ref{fig:analysis_cifar_all}) are displayed in a scatter plot.
    }
    \label{fig:correlations}
\end{figure*}

The correlations between $Dist(F1, F2)$, which is the quantitative measurement of the distance between models' decision boundaries, and the fooled ratio in Figure~\ref{fig:corr_dist_fooled}. We confirmed that the distance metric of decision boundaries $Dist(F1, F2)$ is directly related to the non-target transferability, as stated by Tramer et al.\cite{tramer2017space_of_trans}. In addition, the correlations between $Dist(F1, F2)$ and the same mistake ratio for all evaluated models and adversarial attacks are shown in Figure~\ref{fig:corr_dist_same}. These results show that non-targeted transferability and same mistakes are strongly associated with each other.

\subsection{Perturbation Size Analysis}
\begin{figure*}[p]
    \centering
        \begin{subfigure}[b]{0.85\linewidth}
            \centering
            \includegraphics[width=\linewidth]{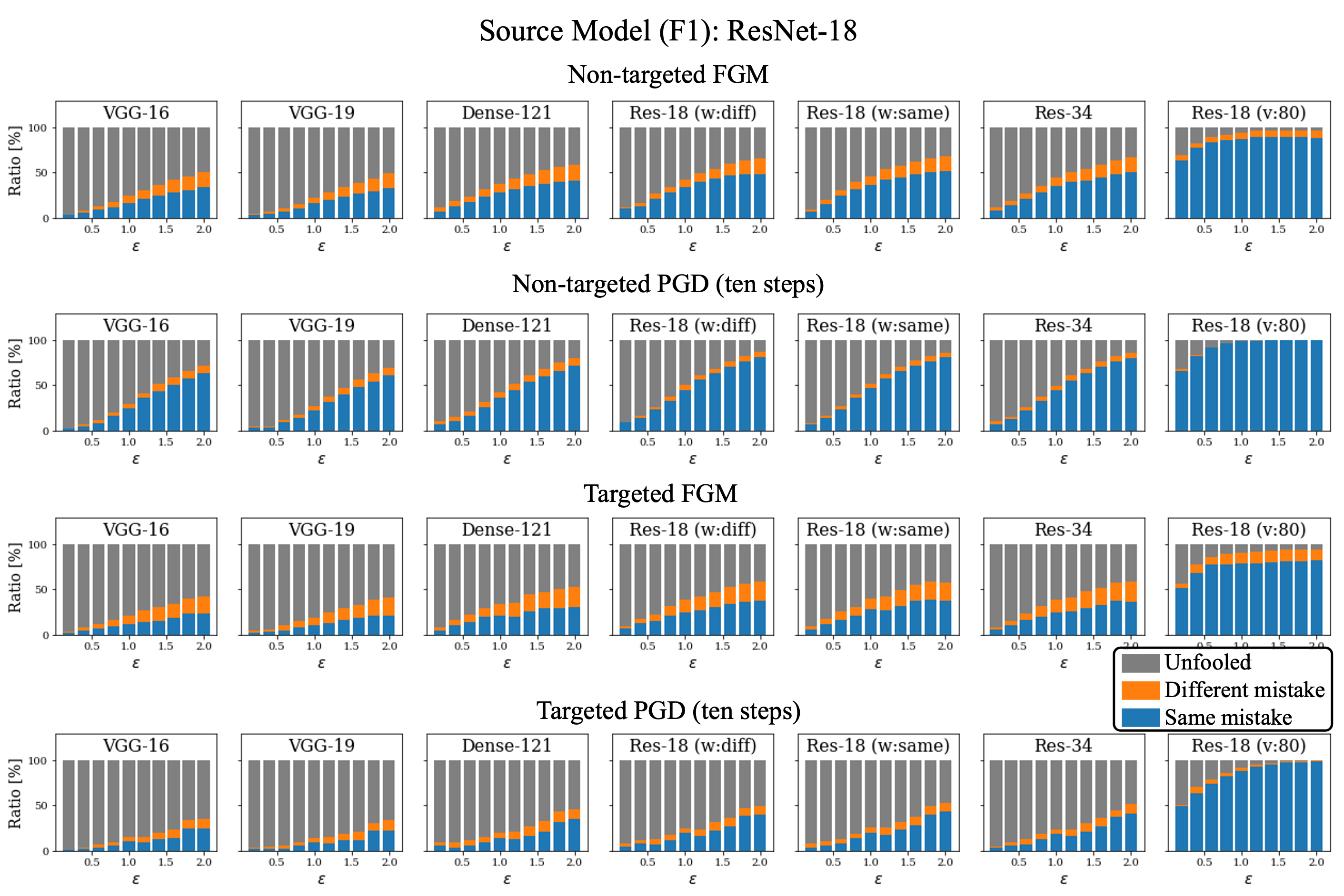} 
            \caption{Class-aware transferability of AEs generated for ResNet-18 source model.}
            \label{fig:eps_res}
        \end{subfigure}
    \quad
        \begin{subfigure}[b]{0.85\linewidth}
            \centering
            \includegraphics[width=\linewidth]{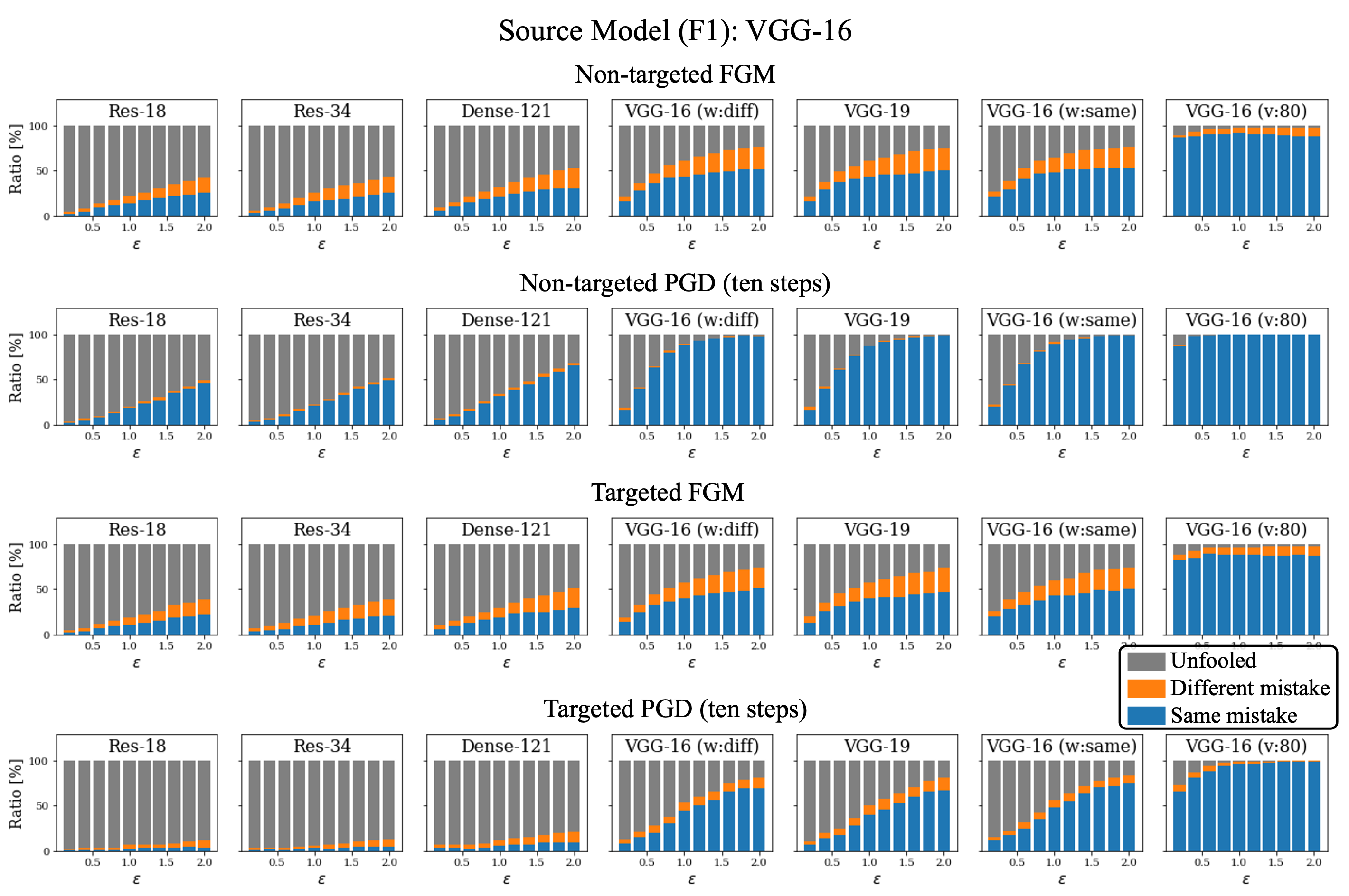}
            \caption{Class-aware transferability of AEs generated for VGG-16 source model.}
            \label{fig:eps_vgg}
        \end{subfigure}
        
    \caption{Class-aware transferability of AEs when the perturbation size $\epsilon$ is gradually changed (CIFAR-10). Here we show the results of attacking the source model of ResNet-18 (Figure~\ref{fig:eps_res}) and VGG-16 (Figure~\ref{fig:eps_vgg}) with non-targeted attack (1st and 2nd rows) and targeted attack (3rd and 4th rows), using FGM (1st and 3rd rows) and PGD (ten-step) (2nd and 4th rows) methods.
    }
    \label{fig:compare_eps_appendix}
\end{figure*}

The class-aware transferability of AEs when the perturbation size was gradually changed is shown in Figure~\ref{fig:eps_res} for the ResNet-18 source model and Figure~\ref{fig:eps_vgg} for the VGG-16 source model.

\subsection{Decision Boundary Analysis}
\begin{figure*}[p]
    \centering
        \includegraphics[width=\linewidth]{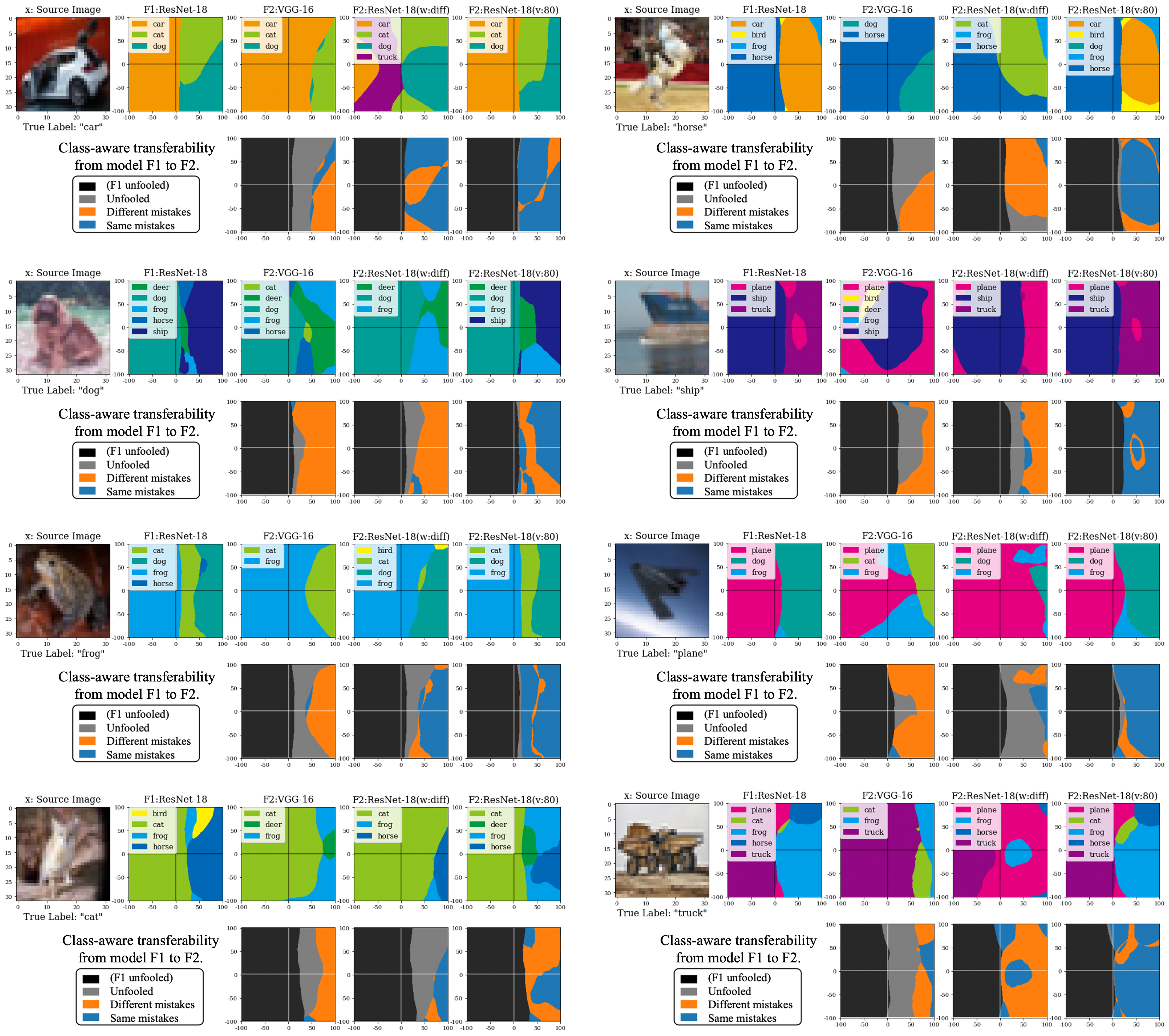} 
        \caption{Visualization of decision boundaries when the source model is ResNet-18 (CIFAR-10). For each image, the first row shows the classification results, where each color represents a certain class. The second row shows to which area the three cases of class-aware transferability correspond. The distance from (0, 0) point to the closest decision boundary along the x-axis corresponds to the metric $d(F1, x)$ for each image $x$. The unit of each axis is 0.02 in l2 distance.
        }
        \label{fig:dec_res}
\end{figure*}

\begin{figure*}[p]
    \centering
        \includegraphics[width=\linewidth]{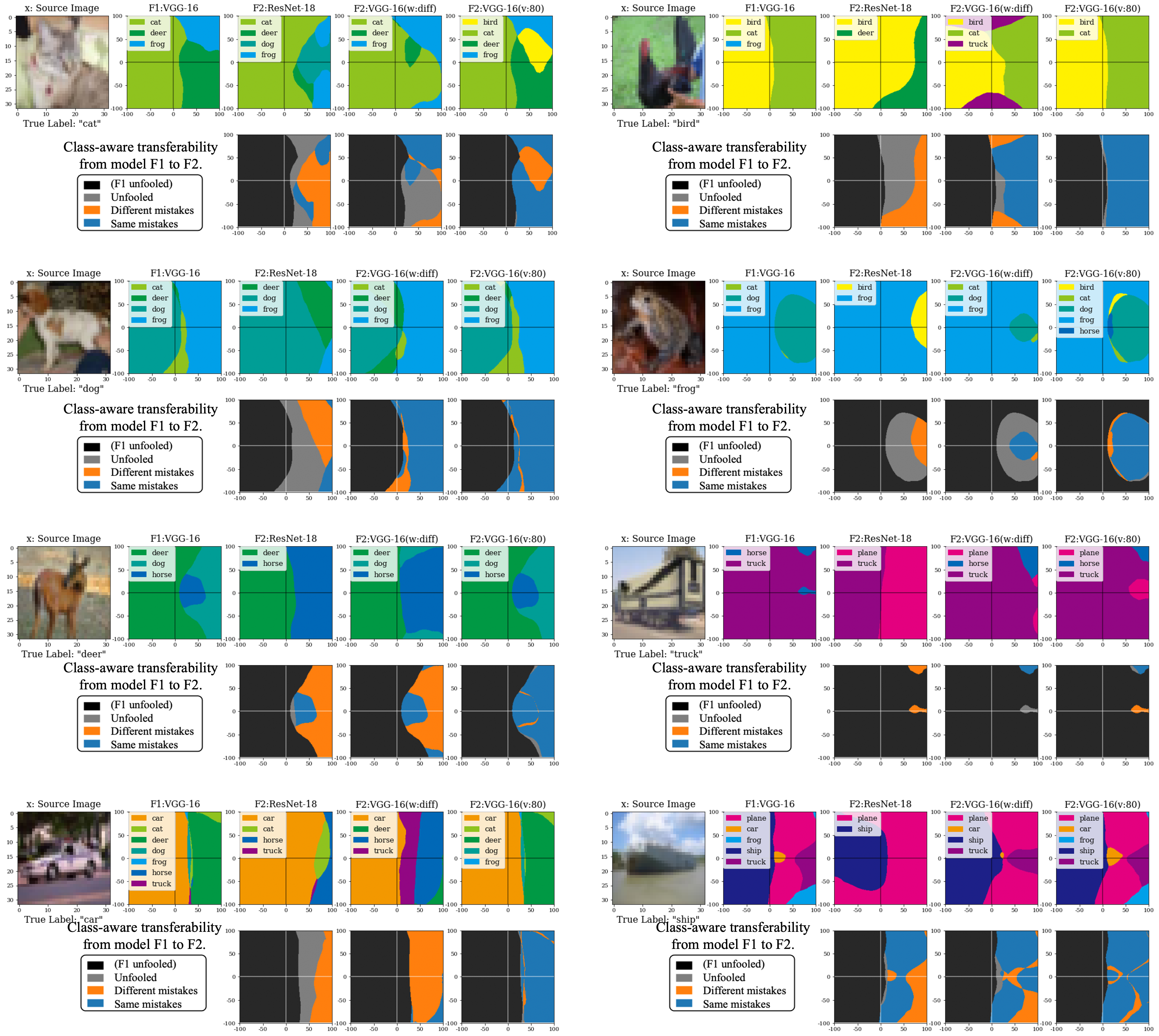} 
        \caption{Visualization of decision boundaries when the source model is VGG-16 (CIFAR-10). For each image, the first row shows the classification results, where each color represents a certain class. The second row shows to which area the three cases of class-aware transferability correspond. 
        The distance from (0, 0) point to the closest decision boundary along the x-axis corresponds to the metric $d(F1, x)$ for each image $x$. 
        The unit of each axis is 0.02 in l2 distance.
        }
        \label{fig:dec_vgg}
\end{figure*}

The visualization of the decision boundary for several different images is shown in Figure~\ref{fig:dec_res} for the ResNet-18 source model and Figure~\ref{fig:dec_vgg} for the VGG-16 source model.

\subsection{Additional Adversarial Transferability Analysis on FGVC-Aircraft dataset}
\vfill
We additionally analyze the class-aware transferability of AEs generated for FGVC-Aircraft dataset \cite{maji2013fine_fgvc_aircraft} to understand the effect of class similarity and the number of classes.
FGVC Aircraft dataset contains 10,000 images, which are split into 6,667 images for train set and 3,333 images for test set. 
It is composed of only images of aircrafts, which are labeled hierarchically: For example, the label level of ``variant", e.g. ``Boeing 737-700", has 100 classes which are finest visually distinguishable classes. 
The label level of ``manufacturer", e.g. ``Boeing", has 40 classes of different manufacturers.
We trained all models at the initial learning rate of $0.01$, which decayed $0.1$ times at the $100^{th}$ and $150^{th}$ epoch with 200 epochs in total.

Since FGVC-Aircraft contains only aircraft images, images for different classes are visually more similar than, e.g., ``cat" and ``truck" images in CIFAR-10. 
Therefore, it is more likely that AEs cause different mistakes unless the AEs have a substantial effect on fooling models towards a specific class.

\begin{figure*}[p]
    \centering
        \includegraphics[width=0.7\linewidth]{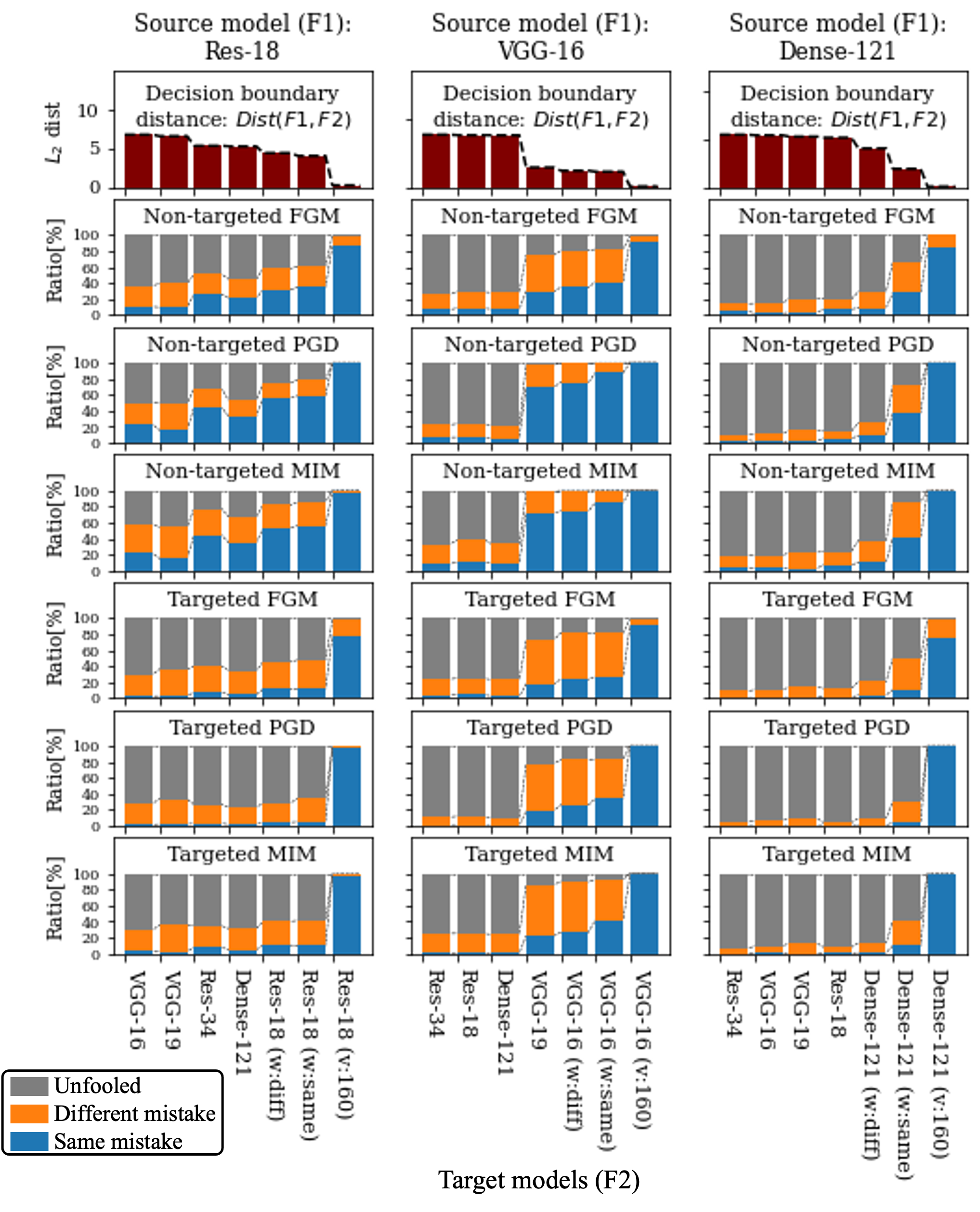} 
        \caption{Class-aware transferability of adversarial attacks for FGVC-Aircraft (``variant"). AEs were $l_2$-bounded by $\epsilon$=5.0.
        Order of F2 is sorted by $Dist(F1, F2)$ (1st row) for each F1 so rightmost F2 was estimated to be more similar to F1.
        }
        \label{fig:analysis_fgvc}
\end{figure*}

Figure~\ref{fig:analysis_fgvc} shows the class-aware transferability of AEs generated for FGVC-Aircraft (``variants") dataset.
Note that if AEs fool target models towards random directions, the proportion of same mistake ratio out of fooled ratio is 1\% for a 100-class dataset.

We observe that non-targeted attacks caused same mistakes at a high rate. 
On the other hand, targeted attacks did not cause same mistakes as many as non-targeted attacks; however, still the proportions of same mistake ratio out of fooled ratio were more than 1\%.

It is intriguing that, although FGVC-Aircraft (``variant") has 100 classes, the same mistake ratio is high with non-targeted attacks.
It indicates that the AEs generated by non-targeted attacks had strong effects on fooling models towards specific classes, which suggests the existence of non-robust features of the specific classes.

For targeted attacks, although targeted attacks still cause a moderate number of same mistakes, they are not as much as non-targeted attacks. 
For example, the proportions of same mistake ratio out of fooled ratio when \{F1, F2\}=\{ResNet-18,VGG-16\} were 9.4\%, 6.0\%, and 9.8\% for targeted FGM, PGD, and MIM, respectively (the leftmost column of Figure~\ref{fig:analysis_fgvc}).

\begin{figure*}[p]
    \centering
        \begin{subfigure}[b]{0.8\linewidth}
            \centering
            \includegraphics[width=0.66\linewidth]{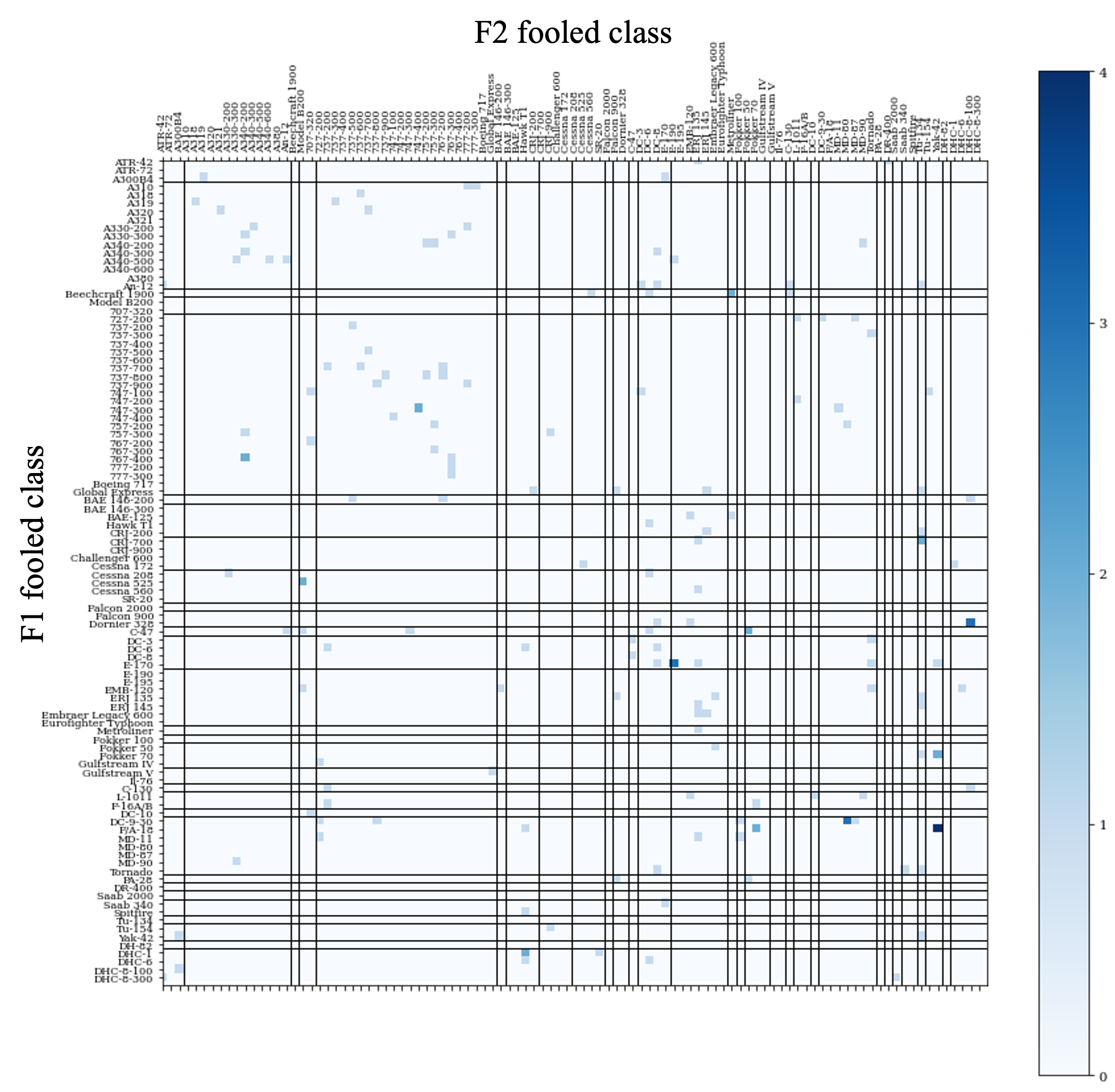} 
            \caption{Class-wise analysis of different mistakes caused by AEs generated by non-targeted FGM.
            }
            \label{fig:class_wise_non_tar}
        \end{subfigure}
    \quad
        \begin{subfigure}[b]{0.8\linewidth}
            \centering
            \includegraphics[width=0.66\linewidth]{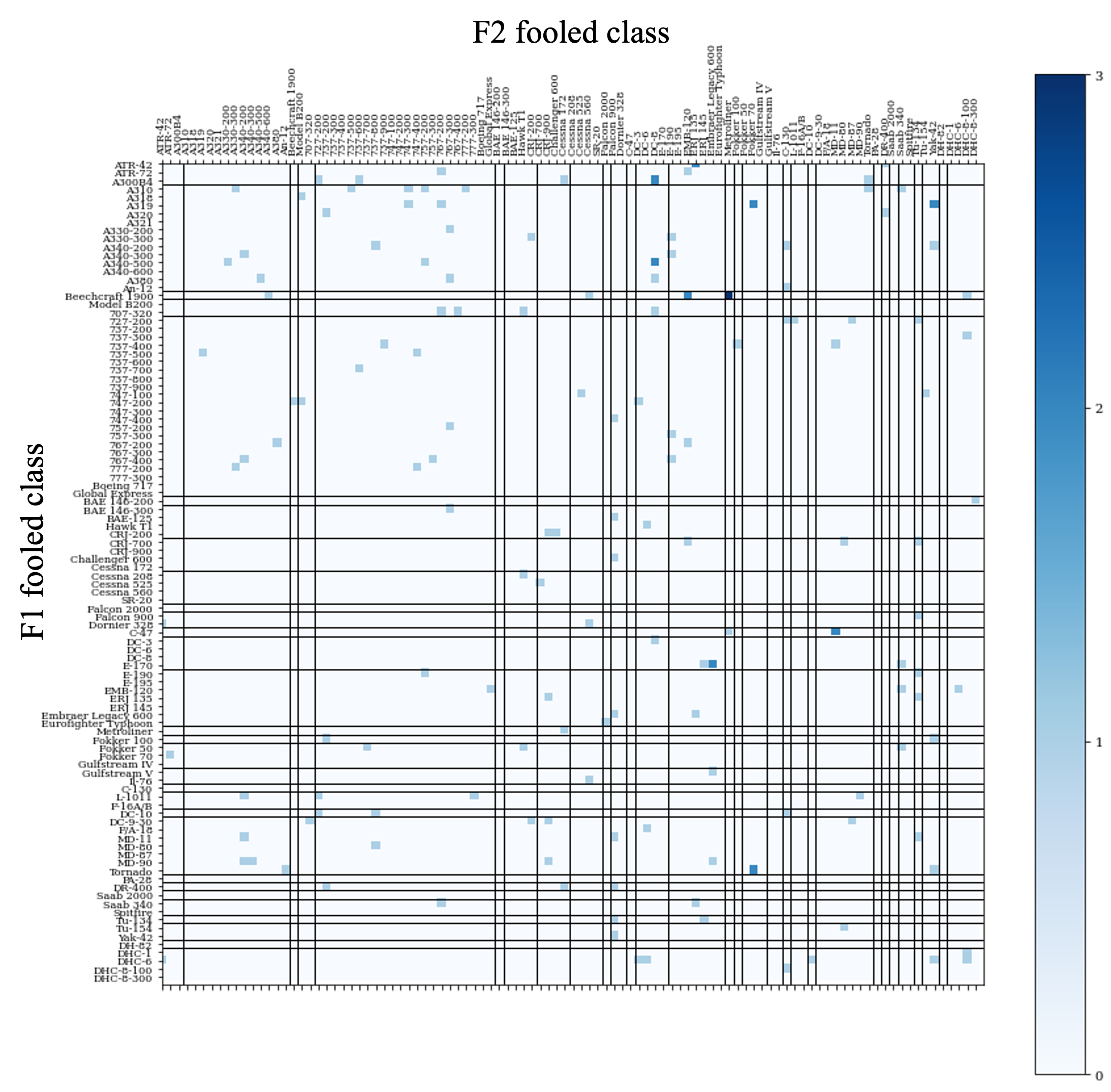}
            \caption{Class-wise analysis of different mistakes caused by AEs generated by targeted FGM.
            }
            \label{fig:class_wise_tar}
        \end{subfigure}
        
    \caption{Class-wise analysis of ``different mistakes" for FGVC-Aircraft (``variant"). 
    The y-axis shows the classes to which the source model F1 misclassified the AEs and the x-axis shows the classes to which the target model F2 misclassified the AEs.
    Each value represents the number of each case.
    The source model F1 is ResNet-18 and the target model F2 is VGG-16.
    The classes are sorted by the ``manufacturer" labels and the black lines separates ``variant" classes for each ``manufacturer".
    For non-targeted FGM (Figure~\ref{fig:class_wise_non_tar}), it is observed that different mistakes tend to occur within the same ``manufacturer".
    On the other hand, targeted FGM (Figure~\ref{fig:class_wise_tar}) caused different mistakes for other ``manufacturers" more than non-targeted FGM.
    }
    \label{fig:class_wise_fgvc}
\end{figure*}

To understand how different mistakes occur with the FGVC-Aircraft dataset, we further analyzed different mistakes at a class-wise level (Figure~\ref{fig:class_wise_fgvc}).
For non-targeted FGM (Figure~\ref{fig:class_wise_non_tar}), it is observed that different mistakes tend to occur within the same ``manufacturer".
It indicates that in the different mistake cases in non-targeted AEs, the non-robust features of a specific class were recognized as a different but similar class.
On the other hand, targeted FGM (Figure~\ref{fig:class_wise_tar}) caused different mistakes for other ``manufacturers" more than non-targeted FGM.
In general, targeted attacks are harder to perform than non-targeted attacks since targeted attacks are forced to aim at a specific class.
Therefore, we think that this difficulty of targeted attacks can result in targeted attacks generating AEs with model-specific non-robust features for the source model, which are not likely to be perceived similarly by a target model.
However, the differences between the mechanism and nature of targeted and non-targeted attacks are still not fully understood, which should be future work.

\vspace{0mm}
\section{Non-robust Feature Analysis}
\vspace{0mm}
\subsection{Theory: The Difference in Learned Features Causes Different Behavior on Adversarial Attacks}

In the paper, we showed that different models might classify AEs differently due to the different usage of non-robust features.
In this section, we show a mathematical example of this phenomenon using a simple mathematical model proposed by Tspiras et al. \cite{tsipras2018rob_at_odds}.

\subsubsection{Setup} 
As in Tspiras et al. \cite{tsipras2018rob_at_odds}, we consider a binary classification task in which the data consists of input-label pairs $(x,y)$ sampled from a distribution $D$ as follows: 
\begin{eqnarray*}
    y \overset{u.a.r}{\sim} \{-1, 1\}, \;\;
    x_{1} = 
        \left\{
            \begin{array}{ll}
                +y & w.p. \;\;p \\
                -y & w.p. \;\;1-p
            \end{array}
        \right. , \\
    x_{2},...,x_{d+1} \overset{i.i.d}{\sim} N(\eta y, 1),
\end{eqnarray*}
where $N(\mu, \sigma^2)$ is a normal distribution with mean $\mu$ and variance $\sigma^2$, and $p\leq0.5$. Features $x$ include strongly correlated feature $x_{1}$ and weakly correlated features $x_{2},...,x_{d+1}$ with small coefficient $\eta$. Here, the features $x_{2},...,x_{d+1}$ are non-robust to perturbations with size $\eta$.


\subsubsection{Weakly-correlated features suffice standard classification accuracy}
Although $x_{1},...,x_{d+1}$ only weakly correlate, and each cannot be predictive individually, they can be used to acquire good standard accuracy. As shown in \cite{tsipras2018rob_at_odds}, a simple linear classifier 

\begin{eqnarray*}
    f_{avg} (x) := sign(w_{unif}^{T} x), \\
    \text{where} \;\;  w_{unif} := \left[ 0, \frac{1}{d}, ..., \frac{1}{d} \right]
\end{eqnarray*}
can achieve standard accuracy over 99\% when $\eta \geq 3/\sqrt{d}$ (e.g. if d=1000, $\eta \geq 0.095$). Proof is shown below.
\begin{eqnarray*}
    Pr \left[ f_{avg}(x) = y \right] &=& Pr \left[ sign(w_{unif} x) = y \right] \\
    &=& Pr \left[ \frac{y}{d} \sum_{i=1}^{d} N \left( \eta y, 1 \right) > 0  \right] \\
    &=& Pr \left[ N \left( \eta, \frac{1}{d} \right) > 0 \right] \\
    &>& 99\% \; (\text{when} \;\; \eta \geq 3/\sqrt{d})
\end{eqnarray*}
This means that even when features are weakly correlated, their collection could be predictable enough for classification.

\subsubsection{Different usage of weakly-correlated features can cause different predictions}
Next we think of classifiers $f_{A}$, $f_{B}$ which have weights $w_{A}$, $w_{B}$ as below.
\begin{eqnarray*}
    f_{A} (x) &:=& sign(w_{A}^{T} x), \\
    \text{where} \;\;  w_{A} &:=& \frac{2}{d(d+1)} \left[ 0, 1, 2, ..., d \right] \\
    f_{B} (x) &:=& sign(w_{B}^{T} x), \\
    \qquad\qquad \text{where} \;\;  w_{B} &:=& \frac{2}{d(d+1)} \left[ 0, d,\; d-1, ..., 1 \right]
\end{eqnarray*}
These classifiers only use the weakly-correlated features, but they have a bias on weights, different from $w_{unif}$. 
The difference between these two classifiers is that the preference for using weakly correlated features is the opposite. 
These classifiers achieve a standard accuracy of over 99\% when $\eta \geq  \sqrt{\frac{6(2d+1)}{d(d+1)}}$ (e.g. if d=1000, $\eta \geq 0.11$). The proof for $f_{A}$ is shown below (the same calculation also proves for $f_{B}$).

\footnotesize
\begin{eqnarray*}
    Pr \left[ f_{A}(x) = y \right] &=& Pr \left[ sign(w_{A} x) = y \right] \\
    &=& Pr \left[ \frac{2y}{d(d+1)} \sum_{i=1}^{d} i \cdot N\left(\eta y, 1\right) > 0  \right] \\
    &=& Pr \left[ N\left(\eta \frac{d(d+1)}{2}, \frac{d(d+1)(2d+1)}{6}\right) > 0 \right] \\
    &>& 99\% \; \left(\text{when} \;\; \eta \geq \sqrt{\frac{6(2d+1)}{d(d+1)}} \right)
\end{eqnarray*}
\normalsize
Now we think of an adversarial attack that perturbs each feature $x_{i}$ by a moderate $\epsilon$. For instance, if $\epsilon=2\eta$, adversary can shift each weakly-correlated feature towards $-y$. Here, we consider the case in which only the first half of the weakly-correlated features are perturbed by $\epsilon=2\eta$: we consider perturbed features $x'_{2}, ... ,x'_{k+1}$ are sampled i.i.d. from $N(-\eta y, 1)$, where $k=d/2$ (for simplicity, suppose d is an even number and $d\gg2$). Then the probability of $f_{A}$ correctly predicting $y$ is over 90\% when  $\eta \geq  \sqrt{\frac{6(2d+1)}{d(d+1)}}$ (e.g. if d=1000, $\eta \geq 0.11$).

\footnotesize
\begin{eqnarray*}
\begin{aligned}
    Pr \left[ f_{A}(x') = y \right] &= Pr \left[ sign(w_{A} x') = y \right] \\
    &= Pr \biggl[ \frac{2y}{d(d+1)} \biggl( \sum_{i=1}^{k} i \cdot N\left(- \eta y, 1\right) \\
    &\qquad\qquad\qquad + \sum_{i=k+1}^{2k} i \cdot N\left(\eta y, 1\right) \biggl) > 0  \biggl] \\
    &= Pr \left[ N\left(\eta k^{2}, \frac{d(d+1)(2d+1)}{6}\right) > 0 \right] \\
    &= Pr \left[ N\left(\eta \frac{d^2}{4} \sqrt{\frac{6}{d(d+1)(2d+1)}}, 1\right) > 0 \right] \\
    &> 90\% \; \left(\text{when} \;\; \eta \geq \sqrt{\frac{6(2d+1)}{d(d+1)}}\right)
\end{aligned}
\end{eqnarray*}
\normalsize

In the same way, the probability of $f_{B}$ correctly predicting $y$ is less than 10\% when  $\eta \geq  \sqrt{\frac{6(2d+1)}{d(d+1)}}$ (e.g. if d=1000, $\eta \geq 0.11$).

\footnotesize
\begin{eqnarray*}
    Pr \left[ f_{B}(x') = y \right] &=& Pr \left[ sign(w_{B} x') = y \right] \\
    &=& Pr \left[ N\left(- \eta \frac{d^2}{4} \sqrt{\frac{6}{d(d+1)(2d+1)}}, 1\right) > 0 \right] \\
    &<& 10\% \; \left(\text{when} \;\; \eta \geq \sqrt{\frac{6(2d+1)}{d(d+1)}}\right)
\end{eqnarray*}
\normalsize

Therefore, it is proved that there exists a case in which the perturbed input $x'$ is correctly predicted by $f_{A}$ while incorrectly predicted by $f_{B}$. This analysis shows that how each model puts weights on weakly-correlated features can determine the transferability of adversarial examples. 
Similarly, simply extending this analysis to a multi-class setting can theoretically show that there is a possibility to attack different models to cause different mistakes when the models use features differently.

\subsection{Supplementary Results}

Here, we provide supplementary results and details of the non-robust feature analysis.

\subsubsection{N-targeted attack}

\begin{figure}[H]
    \centering
        \begin{subfigure}[!h]{0.35\linewidth}
            \centering
            \includegraphics[width=1\linewidth]{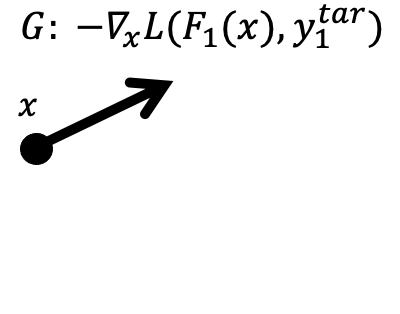} 
            \caption{Targeted attack}
            \label{fig:tar_1}
        \end{subfigure}
    \quad
        \begin{subfigure}[!h]{0.45\linewidth}
            \centering
            \includegraphics[width=1\linewidth]{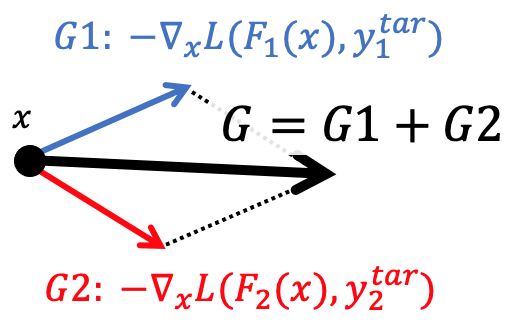}
            \caption{N-targeted attack (N=2)}
            \label{fig:tar_2}
        \end{subfigure}
    \caption{Difference between (a) targeted attack and (b) proposed N-targeted attack (N=2), which sums up all gradients for all target models ($G=G1+G2$) and aims to mislead model $F1$ towards class $y^{tar}_1$ and model $F2$ towards class $y^{tar}_2$.}
    \label{fig:illus_n_tar} 
\end{figure}

Figure~\ref{fig:illus_n_tar} describes the difference between vanilla targeted attack and the N-targeted attack.
N-targeted attack aims to fool multiple models towards each specified target class.
It simply adds up the gradients for all target models. 

Table~\ref{tab:n_tar_accuracy} shows the accuracy of models on the AEs generated by the N-targeted attack, which constructs non-robust sets.

\begin{table*}[p]
\centering
\begin{tabular}{c|cc|c|c|c}
\toprule
    \multirow{2}{*}{Dataset} & \multicolumn{2}{c|}{\begin{tabular}[c]{@{}c@{}}Non-robust set\\ constructed for\end{tabular}} & \multirow{2}{*}{$F1(X')=Y1$} & \multirow{2}{*}{$F2(X')=Y2$} & \multirow{2}{*}{\begin{tabular}[c]{@{}c@{}}$F1(X')=Y1$\\ \& $F2(X')=Y2$\end{tabular}} \\ \cline{2-3}
    & \multicolumn{1}{c|}{F1} & F2 &  &  &  \\ 
\midrule
    \multirow{3}{*}{Fashion-MNIST} & \multicolumn{1}{c|}{Conv-2} & FC-2 & 92.3 & 60.0 & 60.0 \\ \cline{2-6} 
     & \multicolumn{1}{c|}{Conv-2} & Conv-2 (w:same) & 94.6 & 93.8 & 93.0 \\ \cline{2-6} 
     & \multicolumn{1}{c|}{FC-2} & FC-2 (w:same) & 58.7 & 58.5 & 46.0 \\ \hline
    \multirow{3}{*}{CIFAR-10} & \multicolumn{1}{c|}{ResNet-18} & VGG-16 & 95.6 & 99.0 & 95.4 \\ \cline{2-6} 
     & \multicolumn{1}{c|}{ResNet-18} & ResNet-18 (w:same) & 94.1 & 94.1 & 92.0 \\ \cline{2-6} 
     & \multicolumn{1}{c|}{VGG-16} & VGG-16 (w:same) & 99.5 & 99.5 & 99.2 \\ \hline
    \multirow{3}{*}{STL-10} & \multicolumn{1}{c|}{ResNet-18} & VGG-16 & 99.2 & 99.7 & 99.0 \\ \cline{2-6} 
     & \multicolumn{1}{c|}{ResNet-18} & ResNet-18 (w:same) & 99.2 & 99.3 & 99.0 \\ \cline{2-6} 
     & \multicolumn{1}{c|}{VGG-16} & VGG-16 (w:same) & 99.8 & 99.5 & 99.2 \\ 
\bottomrule
\end{tabular}
\caption{Accuracy of models attacked using AEs $X'$ generated by N-targeted attack, which constructs non-robust sets. $Y1$ is target classes for N-targeted attack for model $F1$, and $Y2$ is that for model $F2$. These results are particularly interesting: in a white-box setting, it is easy to generate AEs that lead to different sequences of classes $Y1$ and $Y2$ (success rate $>90\%$ for CIFAR-10 and STL-10).}
\label{tab:n_tar_accuracy}
\end{table*}

\subsubsection{Full Results and Optimized Hyperparameters}
\begin{figure*}[p]
    \centering
        \includegraphics[width=0.7\linewidth]{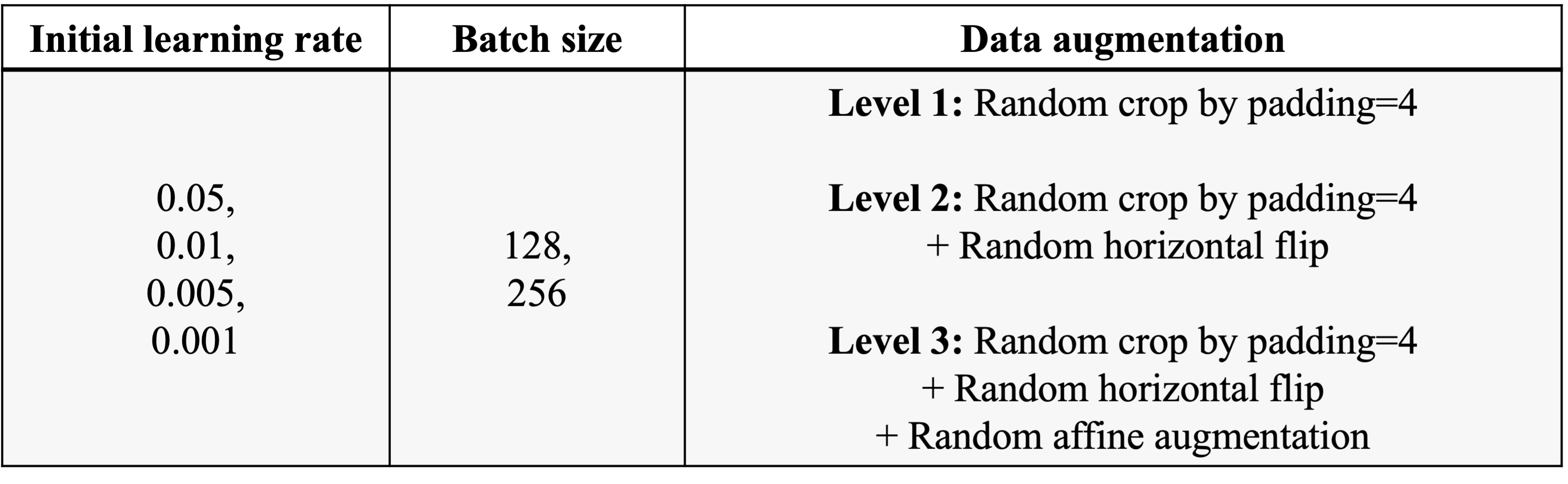} 
        \caption{Grid search area to obtain hyperparameters for training models on the constructed non-robust sets (used for CIFAR-10 and STL-10).}
    \label{fig:grid_search}
\end{figure*}

For CIFAR-10 and STL-10, we conducted a grid search to obtain the best hyperparameters for training models on the constructed non-robust sets. 
The grid search area of hyperparameters is shown in Figure~\ref{fig:grid_search}.
Initial learning rate, batch size, and level of data augmentation were optimized. 
The results and corresponding hyperparameters are shown in Figure~\ref{tab:f_mnist_non_rob_full}, Figure~\ref{tab:cifar10_non_rob_full}, and Figure~\ref{tab:stl10_non_rob_full} for Fashion-MNIST, CIFAR-10, and STL-10, respectively.

\begin{table*}[p]
\centering
\begin{tabular}{c|c|c|c|c||c|c|c}
\toprule
    Dataset & \begin{tabular}[c]{@{}c@{}}Non-robust set\\ constructed for\end{tabular} & \begin{tabular}[c]{@{}c@{}}Train\\ set\end{tabular} & \begin{tabular}[c]{@{}c@{}}Trained\\ model\end{tabular} & \begin{tabular}[c]{@{}c@{}}Test acc\\ (X,Y)\end{tabular} & \begin{tabular}[c]{@{}c@{}}Initial\\ learning\\ rate\end{tabular} & \begin{tabular}[c]{@{}c@{}}Batch \\ size\end{tabular} & \begin{tabular}[c]{@{}c@{}}Data\\ aug.\end{tabular} \\ 
\midrule
    \multirow{12}{*}{\begin{tabular}[c]{@{}c@{}}Fashion-\\ MNIST\end{tabular}} & \multirow{4}{*}{\begin{tabular}[c]{@{}c@{}}F1: Conv-2\\ F2: FC-2\end{tabular}} & \multirow{2}{*}{$D'_1: (X', Y1)$} & Conv-2 & 82.9 & \multirow{12}{*}{0.01} & \multirow{12}{*}{256} & \multirow{12}{*}{None} \\ 
     &  &  & FC-2 & 62.1 &  &  &  \\ \cline{3-5}
     &  & \multirow{2}{*}{$D'_2: (X', Y2)$} & Conv-2 & 80.3 &  &  &  \\ 
     &  &  & FC-2 & 75.4 &  &  &  \\ \cline{2-5}
     & \multirow{4}{*}{\begin{tabular}[c]{@{}c@{}}F1: Conv-2\\ F2: Conv-2\\ (w:same)\end{tabular}} & \multirow{2}{*}{$D'_1: (X', Y1)$} & Conv-2 & 81.9 &  &  &  \\ 
     &  &  & FC-2 & 66.2 &  &  &  \\ \cline{3-5}
     &  & \multirow{2}{*}{$D'_2: (X', Y2)$} & Conv-2 & 82.4 &  &  &  \\ 
     &  &  & FC-2 & 67.1 &  &  &  \\ \cline{2-5}
     & \multirow{4}{*}{\begin{tabular}[c]{@{}c@{}}F1: FC-2\\ F2: FC-2\\ (w:same)\end{tabular}} & \multirow{2}{*}{$D'_1: (X', Y1)$} & Conv-2 & 79.0 &  &  &  \\ 
     &  &  & FC-2 & 80.5 &  &  &  \\ \cline{3-5}
     &  & \multirow{2}{*}{$D'_2: (X', Y2)$} & Conv-2 & 77.6 &  &  &  \\ 
     &  &  & FC-2 & 81.4 &  &  &  \\ 
\bottomrule
\end{tabular}
\caption{Non-robust features analysis for Fashion-MNIST. Initial learning rate, batch size, and data augmentations were fixed.}
\label{tab:f_mnist_non_rob_full}
\end{table*}

\begin{table*}[p]
\centering
\begin{tabular}{c|c|c|c|c||c|c|c}
\toprule
Dataset & \begin{tabular}[c]{@{}c@{}}Non-robust set\\ constructed for\end{tabular} & \begin{tabular}[c]{@{}c@{}}Train\\ set\end{tabular} & \begin{tabular}[c]{@{}c@{}}Trained\\ model\end{tabular} & \begin{tabular}[c]{@{}c@{}}Test acc\\ (X,Y)\end{tabular} & \begin{tabular}[c]{@{}c@{}}Initial\\ learning\\ rate\end{tabular} & \begin{tabular}[c]{@{}c@{}}Batch \\ size\end{tabular} & \begin{tabular}[c]{@{}c@{}}Data\\ aug.\end{tabular} \\ 
\midrule
\multirow{12}{*}{CIFAR-10} & \multirow{4}{*}{\begin{tabular}[c]{@{}c@{}}F1: ResNet-18\\ F2: VGG-16\end{tabular}} & \multirow{2}{*}{$D'_1: (X', Y1)$} & ResNet-18 & 51.3 & 0.005 & 128 & Level 3 \\  
 &  &  & VGG-16\_bn & 53.9 & 0.001 & 128 & Level 2 \\ \cline{3-8} 
 &  & \multirow{2}{*}{$D'_2: (X', Y2)$} & ResNet-18 & 10.2 & 0.05 & 128 & Level 1 \\  
 &  &  & VGG-16\_bn & 71.0 & 0.01 & 128 & Level 1 \\ \cline{2-8} 
 & \multirow{4}{*}{\begin{tabular}[c]{@{}c@{}}F1: ResNet-18\\ F2: ResNet-18\\ (w:same)\end{tabular}} & \multirow{2}{*}{$D'_1: (X', Y1)$} & ResNet-18 & 50.1 & 0.05 & 128 & Level 3 \\  
 &  &  & VGG-16\_bn & 54.1 & 0.005 & 256 & Level 3 \\ \cline{3-8} 
 &  & \multirow{2}{*}{$D'_2: (X', Y2)$} & ResNet-18 & 59.2 & 0.05 & 128 & Level 1 \\  
 &  &  & VGG-16\_bn & 58.9 & 0.005 & 128 & Level 3 \\ \cline{2-8} 
 & \multirow{4}{*}{\begin{tabular}[c]{@{}c@{}}F1: VGG-16\\ F2: VGG-16\\ (w:same)\end{tabular}} & \multirow{2}{*}{$D'_1: (X', Y1)$} & ResNet-18 & 63.5 & 0.05 & 128 & Level 2 \\  
 &  &  & VGG-16 & 68.8 & 0.01 & 256 & Level 3 \\ \cline{3-8} 
 &  & \multirow{2}{*}{$D'_2: (X', Y2)$} & ResNet-18 & 11.0 & 0.01 & 128 & Level 1 \\  
 &  &  & VGG-16 & 73.1 & 0.01 & 128 & Level 1 \\ 
 \bottomrule
\end{tabular}
\caption{Non-robust features analysis for CIFAR-10. 
Optimized hyperparameters are shown besides the test accuracy.}
\label{tab:cifar10_non_rob_full}
\end{table*}

\begin{table*}[p]
\centering
\begin{tabular}{c|c|c|c|c||c|c|c}
\toprule
Dataset & \begin{tabular}[c]{@{}c@{}}Non-robust set\\ constructed for\end{tabular} & \begin{tabular}[c]{@{}c@{}}Train\\ set\end{tabular} & \begin{tabular}[c]{@{}c@{}}Trained\\ model\end{tabular} & \begin{tabular}[c]{@{}c@{}}Test acc\\ (X,Y)\end{tabular} & \begin{tabular}[c]{@{}c@{}}Initial\\ learning\\ rate\end{tabular} & \begin{tabular}[c]{@{}c@{}}Batch \\ size\end{tabular} & \begin{tabular}[c]{@{}c@{}}Data\\ aug.\end{tabular} \\ 
\midrule
\multirow{12}{*}{STL-10} & \multirow{4}{*}{\begin{tabular}[c]{@{}c@{}}F1: ResNet-18\\ F2: VGG-16\end{tabular}} & \multirow{2}{*}{$D'_1: (X', Y1)$} & ResNet-18 & 24.0 & 0.001 & 256 & Level 3 \\  
 &  &  & VGG-16\_bn & 25.4 & 0.001 & 256 & Level 2 \\ \cline{3-8} 
 &  & \multirow{2}{*}{$D'_2: (X', Y2)$} & ResNet-18 & 53.7 & 0.005 & 128 & Level 1 \\  
 &  &  & VGG-16\_bn & 57.2 & 0.01 & 128 & Level 2 \\ \cline{2-8} 
 & \multirow{4}{*}{\begin{tabular}[c]{@{}c@{}}F1: ResNet-18\\ F2: ResNet-18\\ (w:same)\end{tabular}} & \multirow{2}{*}{$D'_1: (X', Y1)$} & ResNet-18 & 18.6 & 0.001 & 256 & Level 3 \\  
 &  &  & VGG-16\_bn & 20.1 & 0.001 & 256 & Level 3 \\ \cline{3-8} 
 &  & \multirow{2}{*}{$D'_2: (X', Y2)$} & ResNet-18 & 32.5 & 0.01 & 128 & Level 1 \\  
 &  &  & VGG-16\_bn & 32.4 & 0.01 & 256 & Level 1 \\ \cline{2-8} 
 & \multirow{4}{*}{\begin{tabular}[c]{@{}c@{}}F1: VGG-16\\ F2: VGG-16\\ (w:same)\end{tabular}} & \multirow{2}{*}{$D'_1: (X', Y1)$} & ResNet-18 & 38.5 & 0.001 & 256 & Level 3 \\  
 &  &  & VGG-16\_bn & 51.8 & 0.01 & 256 & Level 3 \\ \cline{3-8} 
 &  & \multirow{2}{*}{$D'_2: (X', Y2)$} & ResNet-18 & 52.2 & 0.01 & 128 & Level 2 \\  
 &  &  & VGG-16\_bn & 52.2 & 0.01 & 256 & Level 2 \\
\bottomrule
\end{tabular}
\caption{Non-robust features analysis for STL-10. 
Optimized hyperparameters are shown besides the test accuracy.}
\label{tab:stl10_non_rob_full}
\end{table*}

\subsubsection{Accuracy Curves}
\begin{figure*}[p]
    \centering
        \includegraphics[width=\linewidth]{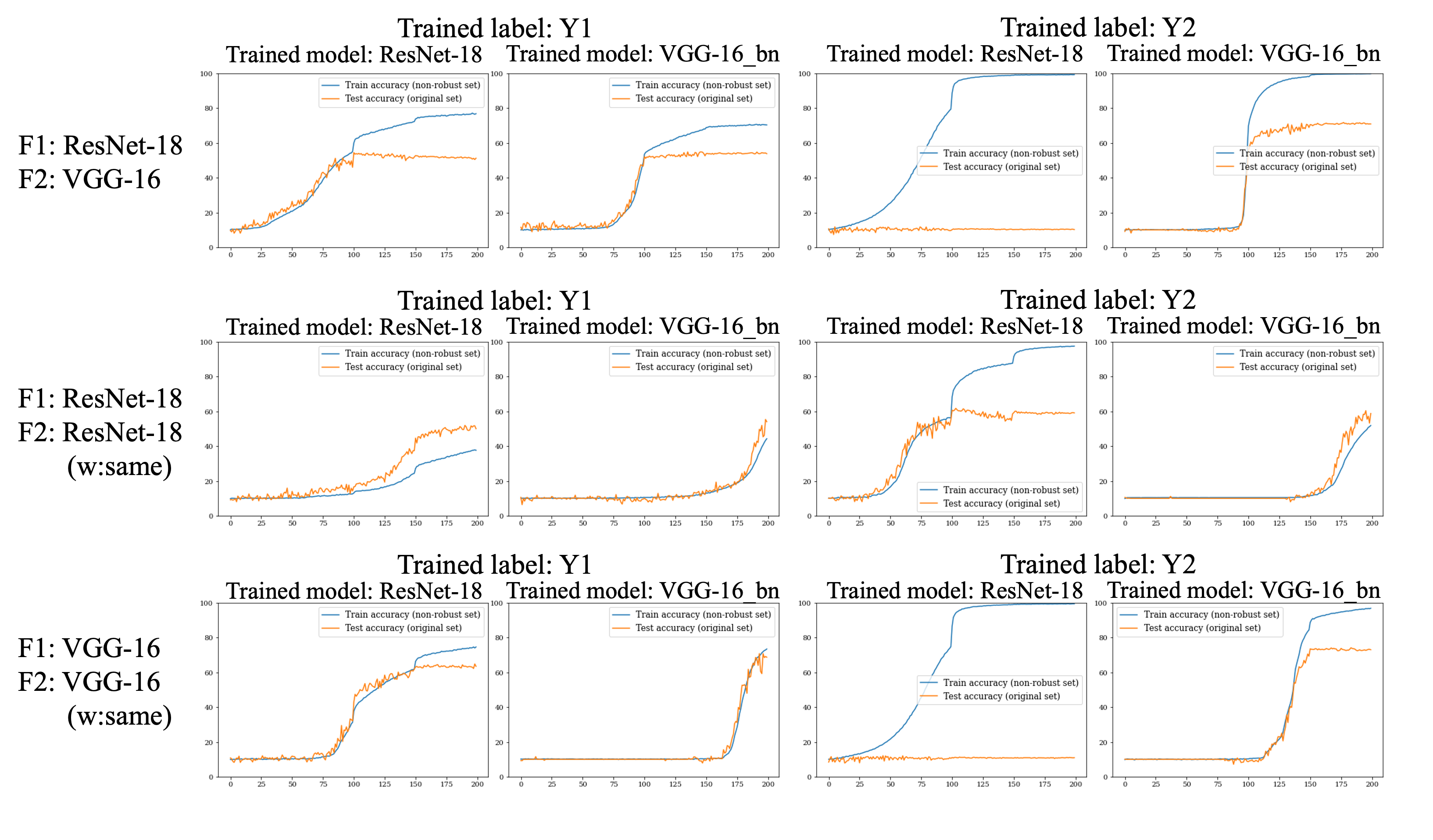} 
        \caption{Accuracy curves when models were trained on the constructed non-robust sets (CIFAR-10). 
        Each figure plots training accuracy on the constructed non-robust set (blue line) and test accuracy on the original test set (orange line).
        }
        \label{fig:acc_curve}
\end{figure*}

Train and test accuracy curves for training models on the constructed non-robust sets are shown in Figure~\ref{fig:acc_curve} (CIFAR-10). 
Note that train accuracy represents accuracy on the constructed non-robust sets, which seem completely mislabeled for humans, and test accuracy represents accuracy on the original test set that is correctly labeled.

Following the experiment from Ilyas et al. \cite{ilyas2019AE_non_rob}, the accuracy numbers reported correspond to the last iteration since we cannot do meaningful early-stopping as the validation set itself comes from the constructed non-robust set and not from the true data distribution.

\section{Potential Application of Our Findings}

In this paper, we have mainly focused on the theoretical understanding of adversarial transferability. This section lists some potential applications to use our main findings.

\subsection{Attack-side perspective}
Using the N-targeted attack concept is one potential application. It can be used to attack systems with the primary classifier model and an AE detection model. Experiments showed that it might be possible to generate AEs with non-robust features that are recognized by the primary classifier but not by the AE detection model. 
Another potential application of our paper is to generate transferable AEs. Our paper suggests that AEs transfer when they have non-robust features that DNNs commonly recognize. Therefore, the promising direction to generate transferable AEs is to investigate how to find ``commonly perceived" non-robust features by different DNNs.

\subsection{Defense-side perspective}
In general, our work further supports viewing adversarial vulnerability as a feature learning problem, as asserted by Ilyas et al. \cite{ilyas2019AE_non_rob}: to reduce the adversarial vulnerability of DNNs, it is necessary to restrict DNNs from learning non-robust features that humans do not use. 
Our contribution is to support this view by showing that non-robust features can explain the transferability of AEs, even from the more detailed perspective of class-aware transferability. 
One specific approach our paper suggests is to ensemble models: it can alleviate the sensitivity to non-robust features learned by a particular model and become only sensitive to the non-robust features commonly learned by all models to be ensembled. 
In other words, the ensemble model may rely less on specific non-robust features than a single model, which can reduce adversarial vulnerability.



\end{document}